\documentclass[letterpaper, 10 pt, journal, twoside]{IEEEtran}

\IEEEoverridecommandlockouts                  
\usepackage{graphics} 
\usepackage{mathptmx} 
\usepackage{amsmath} 
\usepackage{amssymb}  
\usepackage[export]{adjustbox} 
\usepackage{hyperref}
\usepackage{cleveref}
\usepackage{bm}
\usepackage{subcaption} 
\usepackage{tikz}
\usepackage[compact]{titlesec}

\ifCLASSINFOpdf
\else
\fi

\begin{document}
\title{RigidFusion: Robot Localisation and Mapping in Environments with Large Dynamic Rigid Objects
}

\author{Ran Long$^{1,*}$, Christian Rauch$^{1}$, Tianwei Zhang$^{2}$, Vladimir Ivan$^{1}$, Sethu Vijayakumar$^{1,3,*}$%
\thanks{Manuscript received: October 15, 2020; Revised January 12, 2021; Accepted February 16, 2021.}
\thanks{This paper was recommended for publication by Associate Editor J. Civera and Editor A. M. Okamura upon evaluation of the Reviewers' comments.}
\thanks{This research is supported by EPSRC UK RAI Hub for Offshore Robotics for Certification of Assets (ORCA, EP/R026173/1), The Alan Turing Institute and Shenzhen Institute of Artificial Intelligence and Robotics for Society.}
\thanks{$^{1}$Ran Long, Christian Rauch, Vladimir Ivan and Sethu Vijayakumar are with the Institute of Perception, Action and Behaviour, School of Informatics, University of Edinburgh, Edinburgh, EH8 9AB, U.K.}%
\thanks{$^{2}$Tianwei Zhang is with the Shenzhen Institute of Artificial Intelligence and Robotics for Society (AIRS).}%
\thanks{$^{3}$Sethu Vijayakumar is a visiting researcher at the Shenzhen Institute for Artificial Intelligence and Robotics for Society (AIRS).}
\thanks{$^{*}$Corresponding authors: Ran Long (\texttt{Ran.Long@ed.ac.uk}), Sethu Vijayakumar (\texttt{Sethu.Vijayakumar@ed.ac.uk}).
}%
\thanks{Digital Object Identifier (DOI): see top of this page.}%
}

\markboth{IEEE Robotics and Automation Letters. Preprint Version. Accepted February, 2021}
{Long \MakeLowercase{\textit{et al.}}: RigidFusion} 

\maketitle
\begin{abstract}
This work presents a novel RGB-D SLAM approach to simultaneously segment, track and reconstruct the static background and large dynamic rigid objects that can occlude major portions of the camera view. Previous approaches treat dynamic parts of a scene as outliers and are thus limited to a small amount of changes in the scene, or rely on prior information for all objects in the scene to enable robust camera tracking. Here, we propose to treat all dynamic parts as one rigid body and simultaneously segment and track both static and dynamic components. We, therefore, enable simultaneous localisation and reconstruction of both the static background and rigid dynamic components in environments where dynamic objects cause large occlusion. We evaluate our approach on multiple challenging scenes with large dynamic occlusion. The evaluation demonstrates that our approach achieves better motion segmentation, localisation and mapping without requiring prior knowledge of the dynamic object's shape and appearance.

\end{abstract}


\section{\textsc{Introduction}}
\IEEEPARstart{M}{obile} manipulation tasks, such as handling and transporting objects in an unmanned warehouse or collaborative manipulation \cite{stouraitis2020online}, require a robot to localise against the static environment in which it moves while being robust to distractions from dynamic objects; as well as track the object which they manipulate. While these two problems have been previously addressed separately, only a few strands of work \cite{runz2017co, judd2018multimotion} have attempted to solve these two problems together and track the camera and multiple objects at once.

In this work, we argue that localisation against the environment and tracking of objects are fundamentally the same problem, and that solving them simultaneously reduces ambiguity about the scene and improves localisation in cases of large dynamic occlusions.

\begin{figure}[tbh]
    \centering
    \setlength{\belowcaptionskip}{-0.1cm}
    \setlength{\tabcolsep}{0pt}
    \begin{tabular}{ccccc}
    \rotatebox[origin=c]{90}{RGB}\hspace{0.1cm} & \includegraphics[width=0.235\linewidth,frame,valign=m]{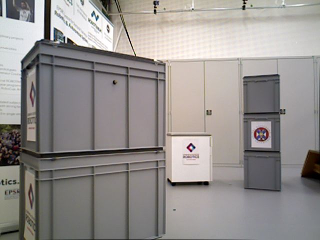}        & \includegraphics[width=0.235\linewidth,frame,valign=m]{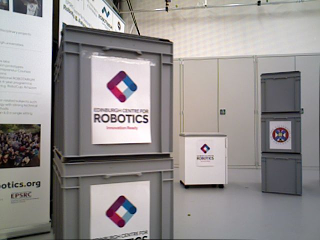}        & \includegraphics[width=0.235\linewidth,frame,valign=m]{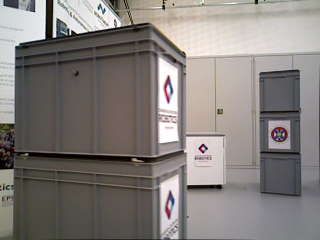}        & \includegraphics[width=0.235\linewidth,frame,valign=m]{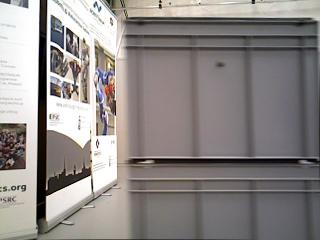}        \\
    \rotatebox[origin=c]{90}{SF}\hspace{0.1cm}  & \includegraphics[width=0.235\linewidth,frame,valign=m]{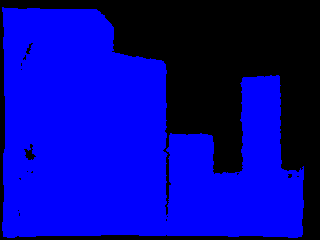} & \includegraphics[width=0.235\linewidth,frame,valign=m]{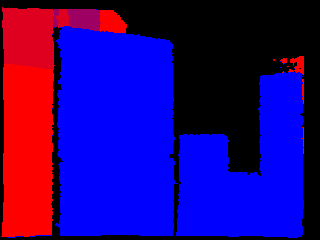} & \includegraphics[width=0.235\linewidth,frame,valign=m]{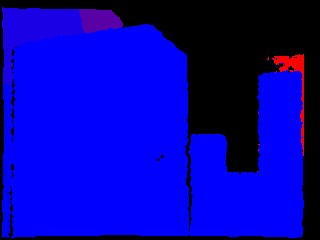} & \includegraphics[width=0.235\linewidth,frame,valign=m]{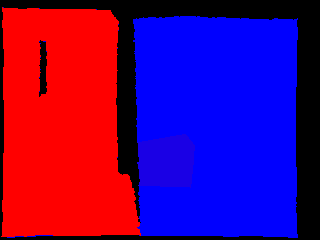} \\
    \rotatebox[origin=c]{90}{RF (ours)}\hspace{0.1cm}  & \includegraphics[width=0.235\linewidth,frame,valign=m]{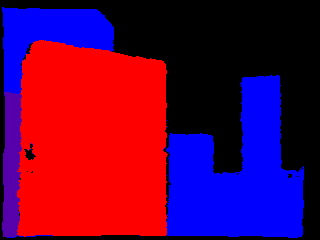}    & \includegraphics[width=0.235\linewidth,frame,valign=m]{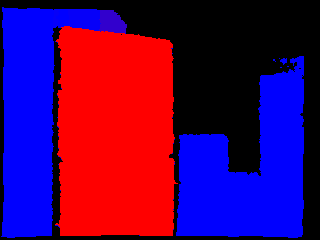}    & \includegraphics[width=0.235\linewidth,frame,valign=m]{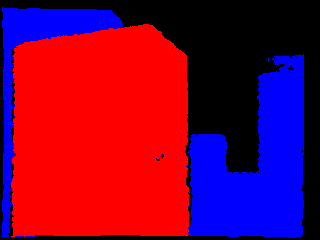}    & \includegraphics[width=0.235\linewidth,frame,valign=m]{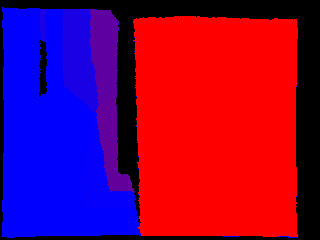}   
    \end{tabular}
    
    \vspace{0.1cm}
    \hrulefill
    \vspace{0.1cm}
    
    \begin{tabular}{c|c}
    \includegraphics[width=0.49\linewidth]{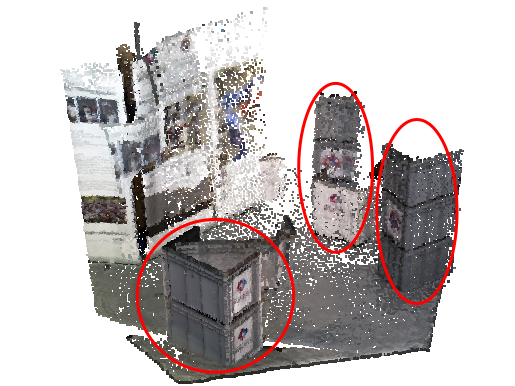} \hspace{0.1cm} & \hspace{0.1cm} \includegraphics[width=0.49\linewidth]{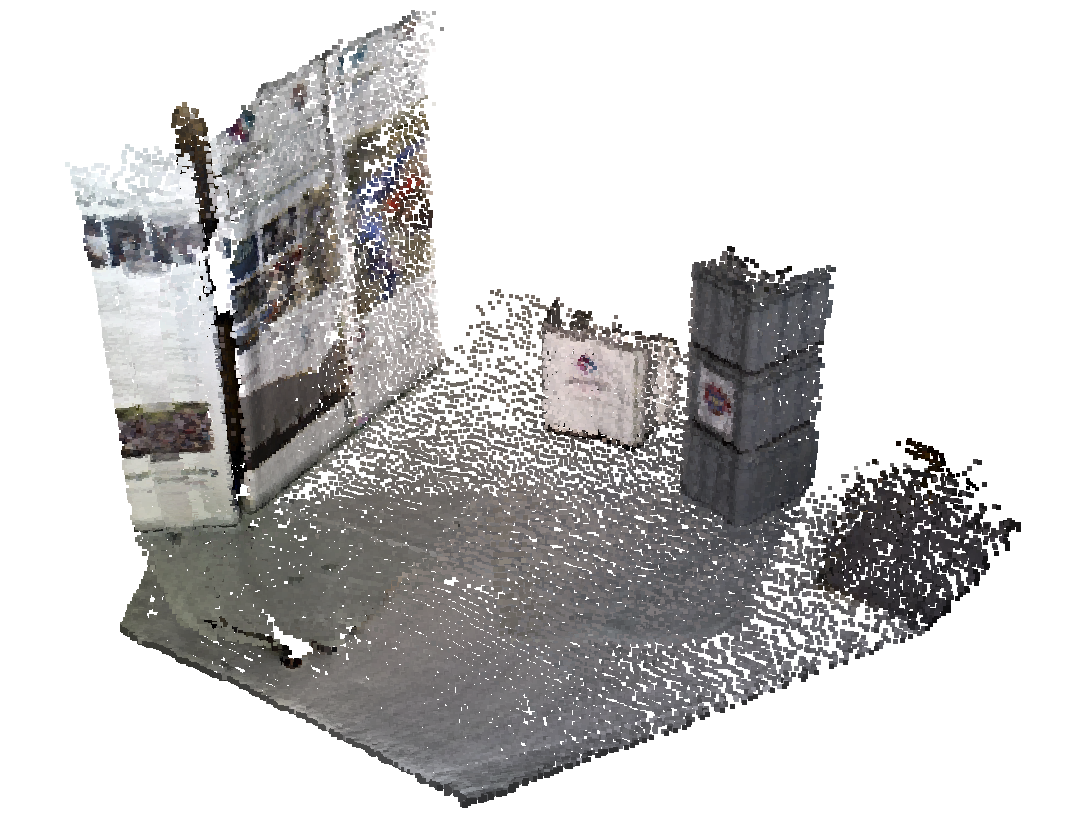} \\
    StaticFusion & RigidFusion 
    \end{tabular}
    
    \caption{\textbf{Top:} Segmentation of a scene with one moving box into static (blue) and dynamic (red) segments.
    Indirect methods, such as StaticFusion (SF) \cite{scona2018staticfusion}, neglect dynamic parts or incorrectly treat them as static environment while our method, RigidFusion (RF), correctly segments the moving box as dynamic (red).
     \textbf{Bottom:} The reconstruction of the static map in SF contains the dynamic object (red circle) and multiple instances of the same static object (red ellipses), while RF correctly incorporates all static segments.}
    \label{fig:mapping}
\end{figure}
The core problem -- separating the scene into segments of transformations induced by ego-motion and/or moving objects -- is challenging due to several factors:

\begin{enumerate}
    \item \textbf{Unknown environments}: Robots may not have prior information about the semantic meaning, 3D model or appearance of the dynamic objects and the background.
    
    \item \textbf{Large occlusion}: Dynamic parts of images are often discarded for robust visual odometry; but in many scenarios, they can occlude the majority of a camera view, such as for large moving objects or when manipulated objects are close to the camera. This ambiguity leads to tracking failures where a dynamic object is classified as part of the static environment. This is in contrast to driving/flying, where ego-motion effects dominate.
    
    \item \textbf{Mutual static and dynamic transition}: Manipulated objects can transition between static and dynamic with respect to the world at any time during manipulation. Maintaining these state transitions purely with visual odometry can be difficult.
\end{enumerate}

To address all three aspects concurrently, we treat localisation and object tracking as an integrated problem and formalise both as modelling and tracking any rigid movement. Consequently, we achieve improved motion segmentation, localisation and mapping in dynamic environments with large occlusion (\Cref{fig:mapping}). For this, we assume that the motions of both static and dynamic components are rigid transformations. These motions can be identified using tightly coupled motion priors from odometry and kinematics.

In summary, this work contributes:
\begin{enumerate}
    \item a new pipeline to simultaneously segment, track and reconstruct the static background and one dynamic rigid body from RGB-D sequences, using motion priors with potential drift,
    \item a dense SLAM method that is robust to large occlusions in the visual input (over 65\%) without relying on an initialisation of the static and dynamic models,
    \item a new RGB-D SLAM dataset\footnote{\url{http://conferences.inf.ed.ac.uk/rigidfusion/}} with dynamic objects that cause large occlusion in the scenes and ground truth trajectories.
\end{enumerate}

\section{\textsc{Related Work}}
\label{sec:relatedwork}
Dynamic visual SLAM or visual odometry methods can be categorised into \textbf{direct}, \textbf{indirect} or \textbf{multi-motion odometry} methods. Robot proprioception can also be used to support localisation in dynamic environments.

\textbf{Direct methods} rely on prior information of static background or dynamic objects to distinguish between them. For certain dynamic objects, such as humans, PoseFusion \cite{zhang2018posefusion} used OpenPose \cite{openpose} to segment them against the environment. In addition, multi-object segmentation methods, such as Mask-RCNN \cite{he2017mask}, can provide accurate semantic segmentation, therefore supporting robot localisation when dynamic objects are included in the training set \cite{runz2018maskfusion, strecke2019fusion}. Given object segmentation, different objects can be further assigned with different scores to indicate their probability to be dynamic \cite{xiao2019dynamic}. However, a trained network can only recognise objects from the training set, and even if an object is recognised as static, the object can become dynamic if it is manipulated. Another strand of research distinguishes the static background through geometric properties, such as assuming all planes are static \cite{sun2018motion}. This would fail when objects that consist of planes, like boxes, are manipulated.

\textbf{Indirect methods} track the main rigid component of the visual input and discard the remaining components as outliers. Sun et al. \cite{sun2017improving} applied a RANSAC approach to sparse feature points of two consecutive images, and dynamic objects are removed as outliers. This work was later extended to scenarios with multi-cluster dynamics \cite{sun2018motion} and served as a pre-processing step for the input of SLAM algorithms. Li et al. \cite{li2017rgb} proposed a static pixel/point weighting method to represent the probability of a point being static, instead of classifying each point as either absolutely static or dynamic. Both StaticFusion (SF) \cite{scona2018staticfusion} and Joint-VO-SF (JF) \cite{jaimez2017fast} applied a K-Means clustering method to separate the visual input into clusters and similarly assigned static pixel/point weights to each cluster. The dynamic clusters are detected as outliers and SF requires that dynamic components are less than 20-30\% at the initial frame \cite{scona2018staticfusion}. Rather than removing all outliers, Co-Fusion (CF) \cite{runz2017co} treated outliers as an additional object and maintained the model of this object if the outliers 
are connected and occupy more than 3\% of an image. While it maintains multiple objects, it is prone to over-segment the image, thus treating different parts of an object with the same transformation as different entities.

\textbf{Multi-motion odometry} methods, such as multi-body structure from motion (MBSfM) \cite{sabzevari2016multi} and multimotion visual odometry (MVO) \cite{judd2018multimotion}, directly separate and track multiple rigid bodies with distinct motions in the visual input. MBSfM requires all images in advance and cannot be processed online. MVO can also estimate the number of multiple moving objects online based on sparse feature points. However, it cannot provide dense mapping, and the rigid body with the largest number of feature points is treated as static. This means that a dynamic object could be recognised as static if it has a richer texture or occupies a larger part than static background in the visual input.

Robot proprioception can be combined with visual odometry to support localisation. Visual inertial odometry (VIO) methods \cite{leutenegger2015keyframe,qin2018vins} fused IMUs and visual sensors. Wheel \cite{houseago2019ko} or leg \cite{wisth2019robust} odometry can be further combined with VIO to increase the accuracy of localisation. However, they are limited to static environments. Kim et al. \cite{kim2015visual} used an IMU to estimate and, thus, compensate camera ego-motion, therefore removing dynamic objects before camera tracking. However, this method relies on accurate robot proprioception. 

In summary, state-of-the-art visual SLAM methods either 1) require full knowledge of objects in the scene (direct methods) and fail if the dynamic objects are not detected, or 2) cannot handle when dynamic objects become the predominant part of an image (indirect methods).


\section{Overview}
\label{sec:overview}
\begin{figure*}[t]
    \setlength{\belowcaptionskip}{0cm}
    \centering
    \includegraphics[width=\linewidth]{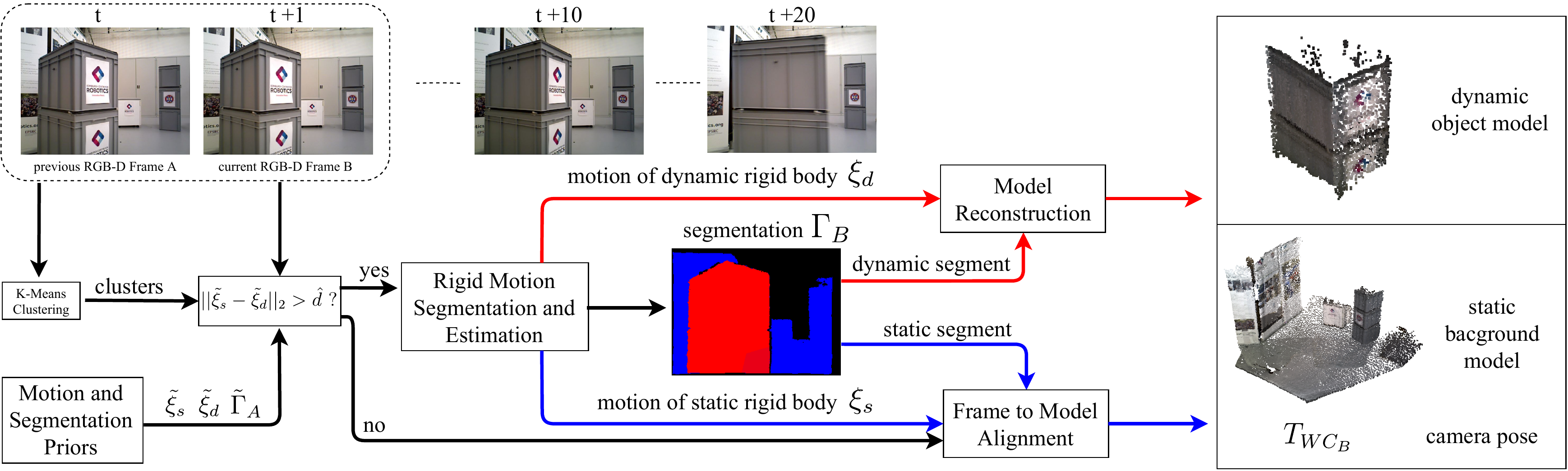}
    \caption{Our method processes two consecutive RGB-D frames (A, B), motion priors ($\tilde{\xi}_s$, $\tilde{\xi}_d$), and the previous segmentation ($\tilde{\Gamma}_A$). We first detect whether the object is dynamic based on motion priors. We then jointly estimate the segmentation $\Gamma_B$ and the rigid body motions $\xi_s$ and $\xi_d$ based on frame-to-frame alignment when the object moves. The segments are used to reconstruct the static environment and the dynamic object, and to localise camera using frame-to-model alignment.}
    \label{fig:pipeline}
\end{figure*}
We propose a pipeline that treats the dynamic component as a single rigid body and uses motion priors to segment the static and dynamic components. The segmentation is used to track the camera, and to reconstruct the background and object models. 

The overview of our pipeline is illustrated in \Cref{fig:pipeline}. Our approach takes two consecutive RGB-D frames A and B, static and dynamic motion priors, $\tilde{\xi}_s, \tilde{\xi}_d \in \mathfrak{se}(3)$, and the previous segmentation of frame A, $\tilde{\Gamma}_A \in \mathbb{R}^{W \times H}$.

Similar to \cite{scona2018staticfusion}, each new intensity and depth image pair $(I,D) \in \mathbb{R}^{W \times H}$ is over-segmented into $K$ geometric clusters $ \mathbf{V} = \{V_i \mid i = 1,\cdots,K\}$ by applying K-Means clustering \cite{jaimez2017fast}. We hypothesise that each cluster is as rigid as possible, and each rigid body can be approximated by the combination of clusters. We also assign each cluster a score $\gamma_i \in [0,1]$ which represents the probability that a cluster belongs to the static rigid body: $\gamma_i = 0$ stands for dynamic clusters while $\gamma_i = 1$ means static clusters. For an RGB-D frame A, we denote the overall scores as $\bm{\gamma}_A  \in \mathbb{R}^K$. 

If the difference between two motion priors $||\tilde{\xi}_s - \tilde{\xi}_d||_2$ is less than a threshold $\hat{d}$, we treat all clusters in the image as static and skip motion segmentation. Otherwise, we jointly optimise the scores $\bm{\gamma}_B$ of the current frame and relative motions $\xi_s$ and $\xi_d$ of the static and dynamic rigid bodies (\Cref{sec:method}).

The pixel-wise segmentation $\Gamma_B \in \mathbb{R}^{W \times H}$ is then computed from clusters and scores. Similar to StaticFusion, we compute the weighted RGB-D images of both static and dynamic rigid bodies from the segmentation $\Gamma_B$. These weighted images are used to reconstruct models of the background and dynamic object and to refine the estimated camera pose using frame-to-model alignment (\Cref{sec:mapping}).

\begin{figure}
    \centering
    \setlength{\belowcaptionskip}{0.2cm}
    \includegraphics[width=\linewidth]{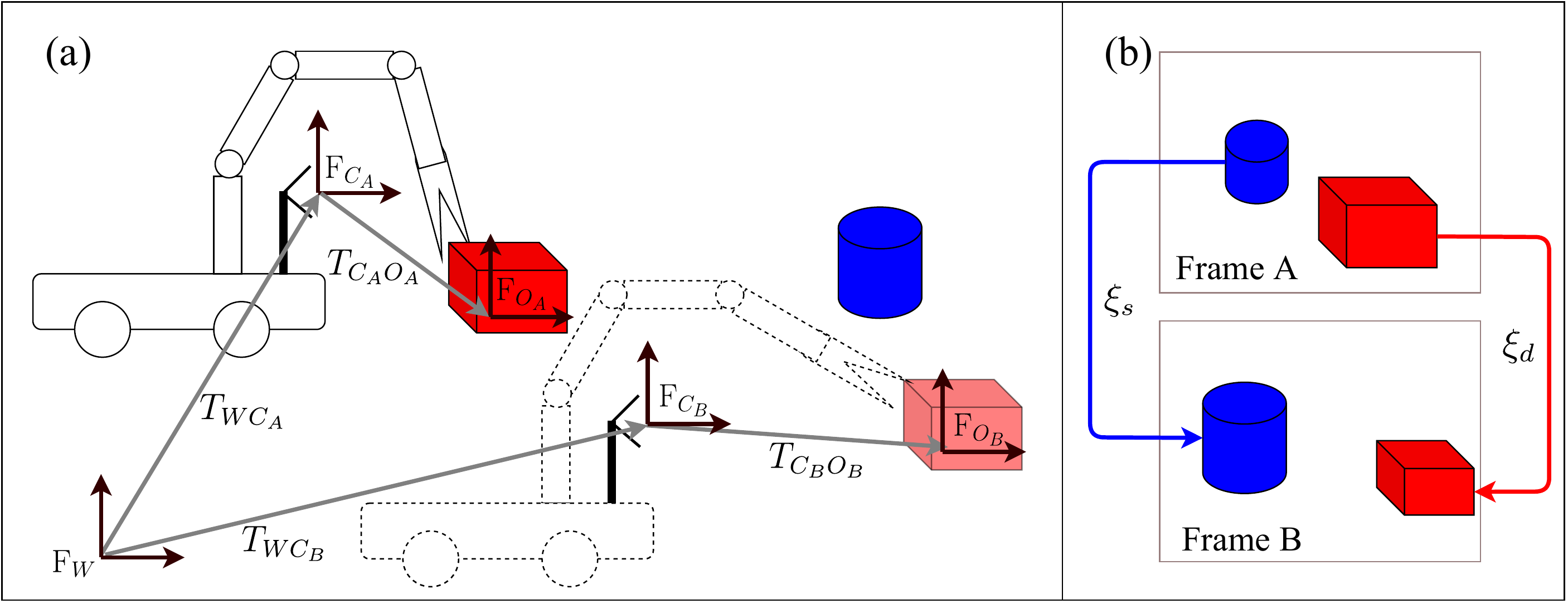}
    \caption{Relation between coordinate frames ($\mathrm{F}_W$, $\mathrm{F}_C$, $\mathrm{F}_O$) and motions ($\xi_s$, $\xi_d$). (a) External camera view. A mobile manipulator simultaneously moves its base and manipulates an object (red box). The camera is fixed on the base. (b) Image view. For the static motion $\xi_s$, we can compute the prior $\tilde{\xi}_s$ from $T_{WC}$, which can be acquired from wheel odometry. The dynamic motion prior $\tilde{\xi}_s$ can be computed from $T_{CO}$, which can be acquired from arm kinematics.}
    \label{fig:frames}
\end{figure}

We denote world-, camera-, and object-frames as $\mathrm{F}_W$, $\mathrm{F}_C$, $\mathrm{F}_O$ respectively (\Cref{fig:frames}). Similar to \cite{houseago2019ko}, we use $T_{XY} \in SE(3)$ to transform homogeneous coordinates of a point in coordinate frame $\mathrm{F}_Y$ to $\mathrm{F}_X$. In an image frame A, the camera and object poses are $T_{WC_A}$ and $T_{WO_A}$ respectively. Considering two image frames A and B, the relation between $\xi_s$ and camera poses is: $T(\xi_s) =T_{WC_A}^{-1}T_{WC_B} = T_{C_AC_B}$, and the relation between $\xi_d$, camera and object poses is: $T(\xi_d) = T_{WC_A}^{-1}T_{WO_A}T_{WO_B}^{-1}T_{WC_B} = T_{C_AO_A}T_{C_BO_B}^{-1}$. The motion priors $\tilde{\xi}_s$ and $\tilde{\xi}_d$ can be provided by proprioceptive sensors, such as wheel odometry and arm forward kinematics. 

In this paper, the static motion prior $\tilde{\xi}_s$ is computed either from wheel odometry or by adding simulated drift on camera ground truth trajectories. We generate $\tilde{\xi}_d$ by simulating drift on object ground truth trajectories.

\section{Rigid Motion Segmentation and Estimation}
\label{sec:method}
At the arrival of each RGB-D pair, we jointly segment and track both static and dynamic rigid bodies by minimising a combined cost that consists of four energy terms:
\begin{equation} \begin{aligned}
    &\min_{\xi_s, \xi_d, \bm{\gamma}} R(\xi_s, \bm{\gamma}) + R(\xi_d, 1-\bm{\gamma}) + S(\xi_d, \bm{\gamma}) + P(\xi_s, \xi_d) \\
    &\text{ s.t.\ $ \gamma_i \in [0,1] \quad \forall \, i$} \ , 
    \label{eq:cost}
\end{aligned} \end{equation}
where $\bm{\gamma}$ represents the scores of all clusters. Specifically, the first two terms align the static and dynamic rigid bodies respectively. The third term $S(\xi_d, \bm{\gamma})$ adds regularisation on both the spatial and time distribution 
of scores $\bm{\gamma}$ to maintain the smoothness of segmentation. The last term $P(\xi_s, \xi_d)$ applies constraints on transformations $\xi_s, \xi_d$ using motion priors $\tilde{\xi}_s$, $\tilde{\xi}_d$.

\subsection{Rigid Body Motion Estimation}
\label{motion}
Following previous RGB-D SLAM methods \cite{whelan2016elasticfusion, scona2018staticfusion}, in static environments, the relative camera pose between two image frames A and B is estimated by minimising the intensity and depth residuals between the RGB-D image pairs ($I_A, D_A$) and ($I_B, D_B$). At a pixel $p$ in frame A, the intensity residuals $r_I^p$ and depth residuals $r_D^p$ with respect to frame B are defined as:
\begin{align}
    r_I^p &= I_B(W(\mathbf{x}_A^p, T(\xi), D_A)) - I_A(\mathbf{x}_A^p)\\
    r_D^p &= D_B(W(\mathbf{x}_A^p, T(\xi), D_A)) - |T(\xi)\pi^{-1}(\mathbf{x}_A^p, D_A(\mathbf{x}_A^p))|_z \ ,
\end{align}
where the image warping function $W$ is given by:
\begin{align}
    W(\mathbf{x}^p, T, D) = \pi(T\pi^{-1}(\mathbf{x}^p, D(\mathbf{x}^p))) \ .
    \label{equ::warp}
\end{align}
$\mathbf{x}^p$ represents the coordinate of pixel $p$ in the 2D image, $|\cdot|_z$ indicates the $z$-coordinate of a 3D point and $D(\mathbf{x}^p)$ is the depth of pixel $p$. The homogeneous transformation matrix $T(\xi) \in SE(3)$ is computed from 
its Lie algebra $\xi \in \mathfrak{se}(3)$. The projection function $\pi: \mathbb{R}^3 \rightarrow \mathbb{R}^2$ projects 3D points onto the image plane using the camera intrinsic matrix.

According to StaticFusion, given the scores $\bm{\gamma}$ of a rigid body, we can estimate the relative motion of this rigid body by applying the scores to weight residuals. Consequently, only pixels that belong to the rigid body have a high contribution: 
\begin{align}
    R(\xi, \bm{\gamma}) = \sum_{p=1}^{N} \gamma_{i(p)}[C(\alpha_I w_{I}^{p}r_{I}^{p}( \xi)) + C(w_{D}^{p}r_{D}^{p}(\xi))] \ ,
    \label{eq:loss_motion}
\end{align}
where $N$ is the number of images pixels with valid depth reading in one image. $i(p)$ indicates the index of the cluster that contains the pixel $p$, and $\gamma_{i(p)}$ represents the probability that this cluster belongs to the rigid body. $\alpha_I$ is a scale parameter to weight photometric residuals so that they are comparable to depth residuals. The parameters $w_I$ and $w_D$ are computed according to the photometric and depth measurement noise. As in \cite{scona2018staticfusion}, we use the Cauchy robust penalty
\begin{align}
    C(r) = \frac{c^2}{2} log(1 + (\frac{r}{c}^2))
    \label{eq:cauchy}
\end{align}
to robustly control the minimisation of residuals, where $c$ is the inflection point of $C(r)$.

The novelty of our approach is that in equation \ref{eq:cost}, we treat the dynamic component as another rigid body with a different motion, where $\bm{\gamma}$ and $1-\bm{\gamma}$ represents the scores of the static and dynamic rigid body respectively. To simultaneously segment and track the two rigid bodies, we further encourage segmentation smoothness and use tightly coupled motion priors. 

\subsection{Segmentation Smoothness}
\label{segmentation}
First, to maintain spatial smoothness, we use the regularisation term used in StaticFusion to penalise the score difference between adjacent clusters:
\begin{align}
    S_{R}(\bm{\gamma}) = \sum_{i=1}^K\sum_{j=i+1}^K E_{ij}(\gamma_{i} - \gamma_{j})^2 \ ,
    \label{equ::sspace}
\end{align}
where $E_{ij}$ is the adjacent map for the cluster set $\mathbf{V}$. $E_{ij} = 1$ if clusters $i$ and $j$ are adjacent in space, otherwise $E_{ij} = 0$.

Furthermore, we consider the physical constraint that pixels that belong to the dynamic rigid body at the previous frame are likely to be dynamic at the current frame. Therefore, we use the segmentation result from the previous frame as segmentation prior to encourage segmentation smoothness over time:
\begin{align}
    S_{T}(\xi_d, \bm{\gamma}) = \sum_{i=1}^K (\gamma_i - \tilde{\gamma}_i(\xi_d))^2 \ ,
    \label{equ::stime}
\end{align}
where $\tilde{\gamma}_i(\xi_d)$ denotes the projection of $\tilde{\gamma}_i$ from the previous frame B to the current frame A via $\xi_d$:
\begin{align}
    \tilde{\gamma}_i(\xi_d) = \sum_{\mathbf{x}_B^p \in V_i} \frac{\tilde{\Gamma}_A(W(\mathbf{x}_B^p, T(\xi_d)^{-1}, D_B))}{|V_i|} \ .
\end{align}
Here, $V_i$ is the $i$-th cluster of the current frame B, and $\tilde{\Gamma}_A$ is the per-pixel segmentation from the previous frame A. The warping function $W$ (equation \ref{equ::warp}) transforms a pixel $p \in V_i$ according to its coordinate in the current image $\mathbf{x}_B^p$ and the estimated motion of rigid body $\xi_d$. $|V_i|$ denotes the number of pixels in $V_i$. 

The spatial and time smoothness (equation \ref{equ::sspace} and \ref{equ::stime}) are combined and weighted by $\lambda_r$:
\begin{align}
    S(\xi_d, \bm{\gamma}) = \lambda_r (S_{R}(\bm{\gamma}) + S_{T}(\xi_d, \bm{\gamma})) \ ,
\end{align}
to represent the smoothness term $S(\xi_d, \bm{\gamma})$ in equation \ref{eq:cost}.

\subsection{Tightly Coupled Motion Prior}
\label{prior_transformation}
Given the motion priors of both static and dynamic rigid bodies $\tilde{\xi}_s$ and $\tilde{\xi}_d$, we add a regularisation term on the motion of each rigid body:
\begin{align}
    P( \xi_s, \xi_d) = \lambda_s||\xi_s - \tilde{\xi}_s||_2^2 + \lambda_d||\xi_d - \tilde{\xi}_d||_2^2 \ ,
    \label{eq:loss_priorpose}
\end{align}
where parameters $\lambda_s$ and $\lambda_d$ weight the regularisation terms. \mbox{$||\cdot||_2^2$} represents the square of the $\mathrm{L}_2$ norm.
Because potential drift and noise in the motion prior could bias the solution, the prior information is neglected if the current estimated state is closer to the prior than a noise-related threshold. To achieve this, $\lambda_s$ and $\lambda_d$ are independently adapted online. Specifically, $\lambda_{s,d} = 1$ if $||\xi_{s,d} - \tilde{\xi}_{s,d}||_2 > \hat{n}$, otherwise, $\lambda_{s,d} = 0$. $\hat{n}$ is a threshold we choose and is related to the noise level of motion priors.

\subsection{Solver}
The solver is based on StaticFusion. Since we directly align images in equation \ref{eq:cost}, the minimisation problem is solved via a coarse-to-fine scheme. We create an image pyramid for each image frame by iteratively down-sampling each image, which reduces the impact of depth noise. The optimisation starts from the coarsest level. The results of intermediate levels are used to initialise the following level.

For each level of the image pyramid, we decouple motions $\xi_s$ and $\xi_d$ from segmentation $\bm{\gamma}$. Specifically, at each iteration, we first fix $\bm{\gamma}$ and optimise $R(\xi_s, \bm{\gamma}) + R(\xi_d, 1-\bm{\gamma}) + P(\xi_s, \xi_d)$ over $\xi_s$ and $\xi_d$. Then $\xi_s$ and $\xi_d$ are fixed, and we optimise $R(\xi_s, \bm{\gamma}) + R(\xi_d, 1-\bm{\gamma}) +S(\xi_d, \bm{\gamma})$ over $\bm{\gamma}$.

\section{Mapping and Frame-to-model Alignment}
\label{sec:mapping}
After the minimisation of equation \ref{eq:cost}, we use the optimal scores $\bm{\gamma}$ and $1 - \bm{\gamma}$ to compute the weighted images for static and dynamic rigid bodies respectively. The weighted images are fused to the model of rigid bodies, and the estimated motions $\xi_s$ and $\xi_d$ are used to initialise the frame-to-model alignment. We use ElasticFusion without loop closure \cite{whelan2016elasticfusion} to build the model and conduct frame-to-model alignment.

\section{\textsc{Evaluation}}
\label{sec:evaluation}

\subsection{Setup}

\begin{figure}[tb]
    \centering
    \setlength{\belowcaptionskip}{0cm}
    \subfloat[Mobile manipulator Ada]{\includegraphics[height=4cm]{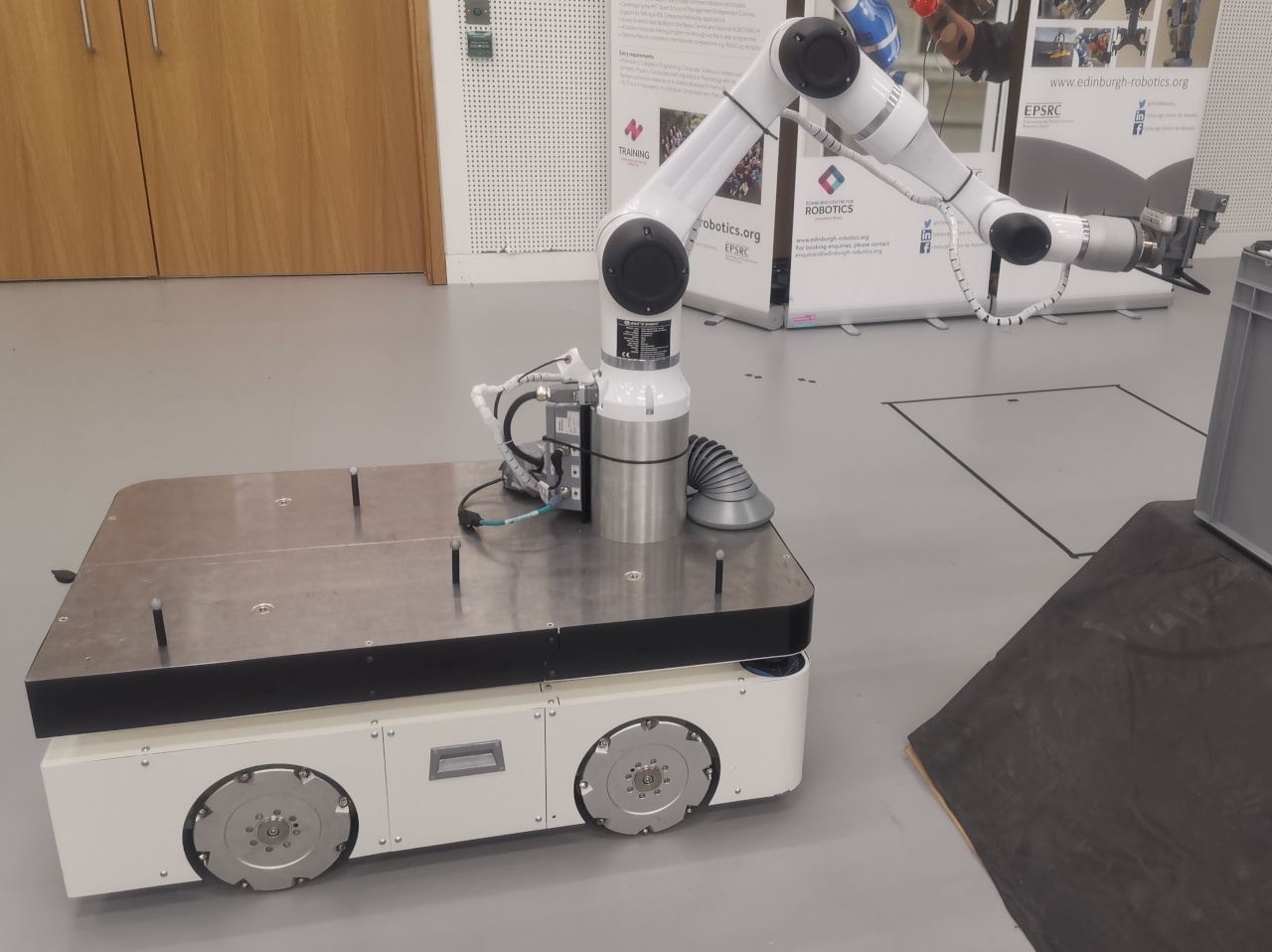} \label{fig:ada}}
    \subfloat[KUKA youBot]{\includegraphics[height=4cm]{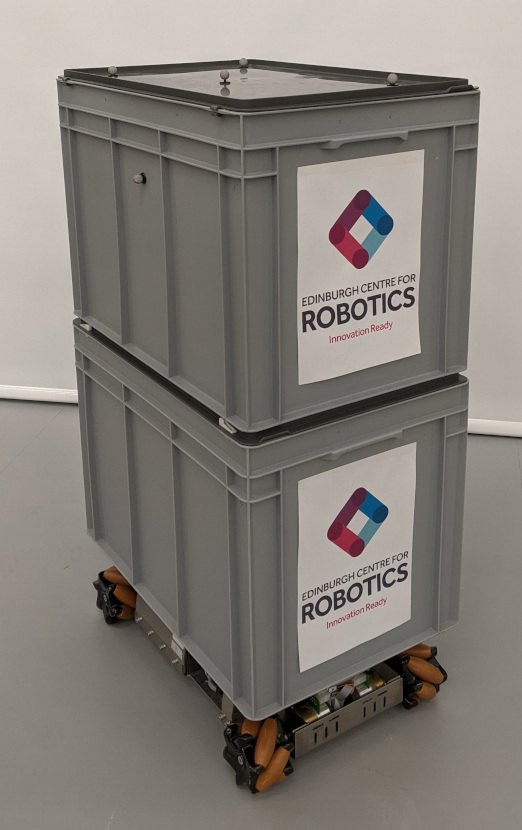} \label{fig:object}}
    \caption{Omnidirectional platforms for moving (\protect\subref{fig:ada}) camera and (\protect\subref{fig:object})  stacked boxes ($0.4 \times 0.6 \times 1$\,m) with Vicon markers.}
    \label{fig:setup}
\end{figure}

The proposed method is evaluated on RGB-D sequences that are collected with an Asus Xtion PRO Live in plane-parallel movement (2 DoF translation. 1 DoF rotation) showing different characteristic object movements. The camera is either hand-held or mounted on an omnidirectional robot base (\Cref{fig:ada}). The object is a remote controlled KUKA youBot with stacked boxes (\Cref{fig:object}). The camera and the object are equipped with Vicon markers for ground truth comparisons and to simulate motion prior drift for camera-only sequences.
The motion estimation performance is quantitatively evaluated via the absolute trajectory error (ATE) and the relative pose error (RPE) \cite{sturm2012benchmark} against the Vicon ground truth for the optical frame. The visualised trajectories are aligned by the initial camera pose.

In the implementation of RF, we set $\lambda_r = 2$, and the thresholds $\hat{d}$ and $\hat{n}$ are both chosen as 0.01. We extend StaticFusion to use motion priors by appending the regularisation term $\lambda_s ||\xi_s - \tilde{\xi}_s||_2^2$ to the loss function. The method that StaticFusion with ground truth camera motion prior is denoted as \emph{SF true}. We control the impact of adding camera motion prior by choosing the same $\hat{n} = 0.01$ for \emph{SF true}. 

For camera-only sequences, the average simulated drift on camera trajectories is 6 cm/s (trans.) and 0.4 rad/s (rot.), while the average drift on object trajectories is 1.5 cm/s (trans.) and 0.1 rad/s (rot.). The camera and object speed is less than 60 cm/s. In robot experiments, we use wheel odometry as camera motion priors and keep the object motion prior with simulated drift.

\subsection{Synthetic Experiments}
\begin{figure}
    \centering
    \setlength{\belowcaptionskip}{0cm}
    \includegraphics[width=\linewidth]{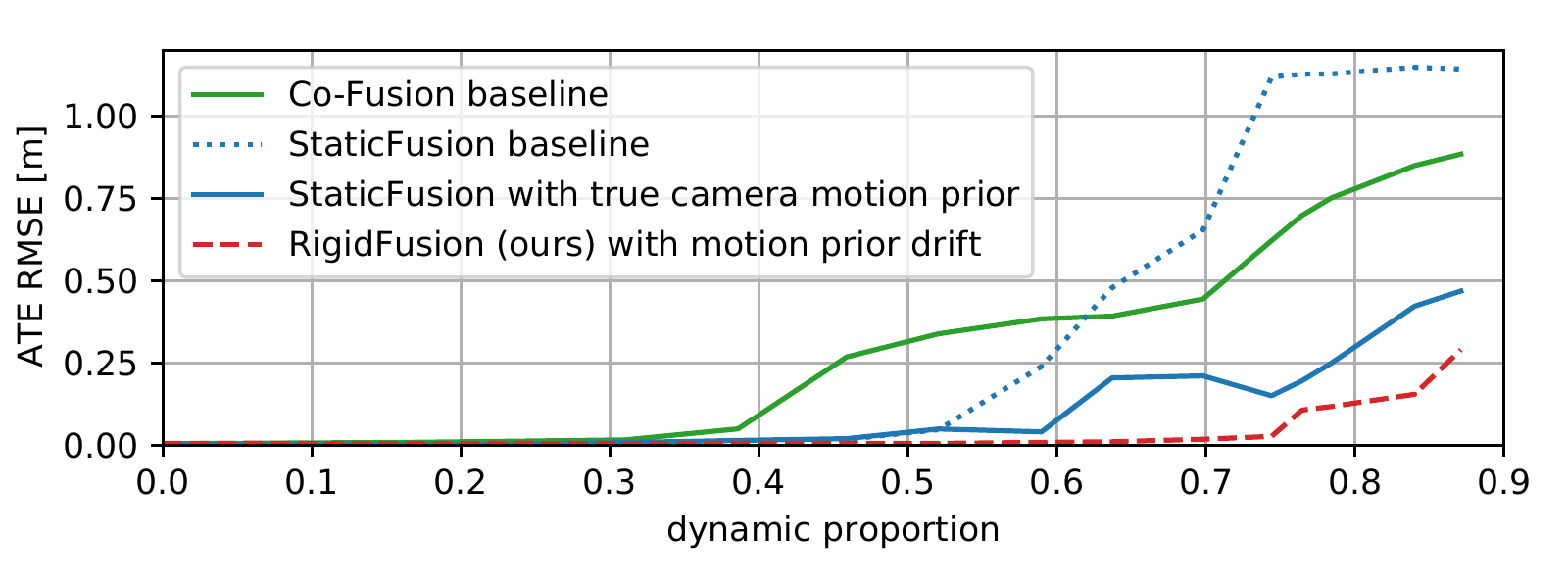}
    \caption{ATE of estimated camera trajectories on a synthetic sequence with different object sizes relative to the amount of valid image pixels. Co-Fusion and StaticFusion break around a dynamic ratio of 0.5 or less. Using the true motion priors in StaticFusion allows larger dynamic objects up to a ratio of 0.6, while our method with drift on the motion priors can track up to a dynamic ratio of 0.75.}
    \label{fig:ablation_size_synth}
\end{figure}

We hypothesise that the proposed objective with motion priors improves the estimation for dynamic objects that occupy more than 50\% of valid image pixels. To study this effect in a controlled environment, we synthesised a simple scene with an object of varying size moving across the image from left to right. The relation of trajectory error to drift magnitude (\Cref{fig:ablation_size_synth}) supports this hypothesis.

\subsection{Camera Experiments}
\label{subsec:camera-only}
\begin{figure*}[tb]
    \centering
    \includegraphics[width=\linewidth]{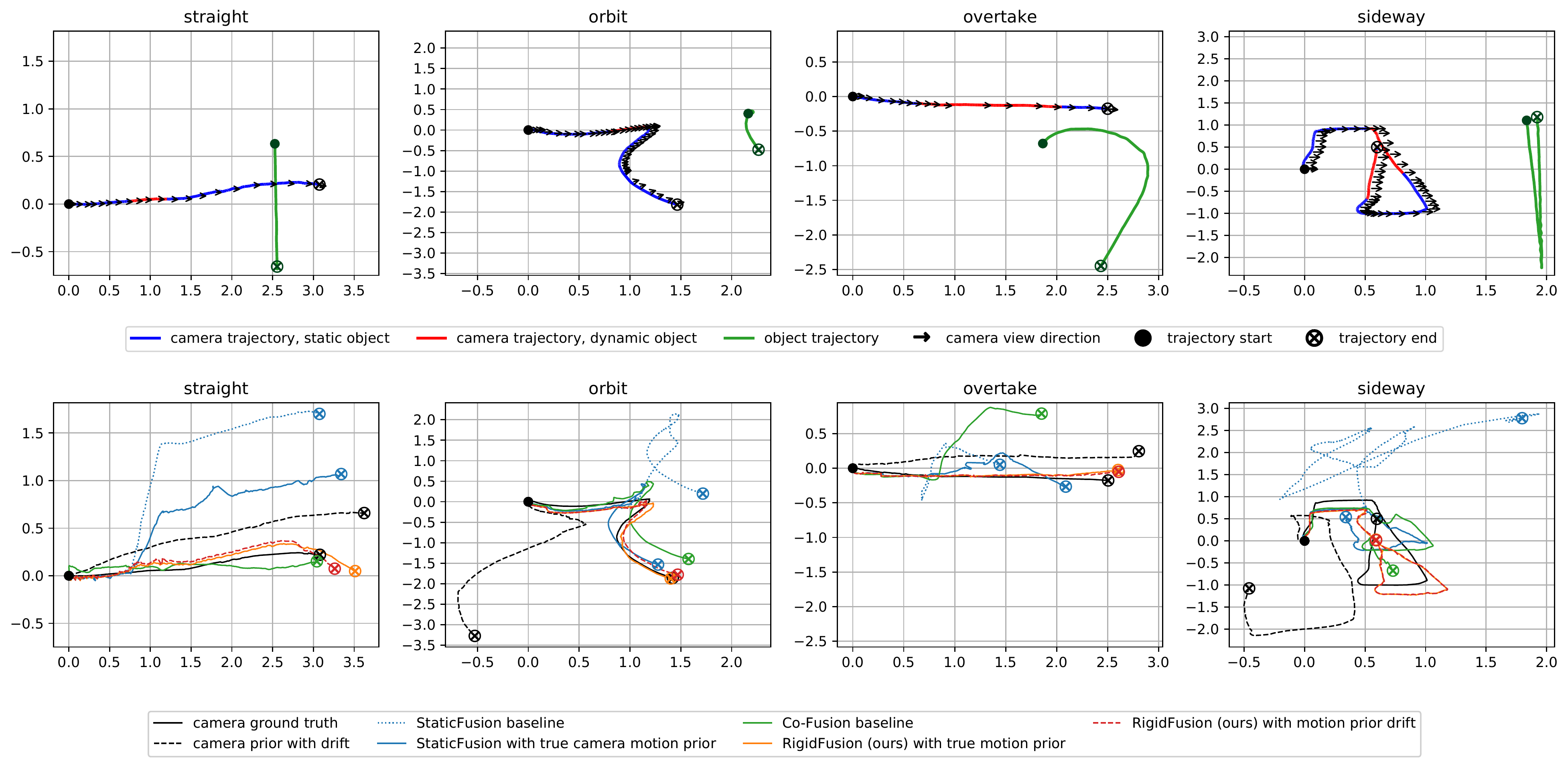}
    \caption{True and estimated trajectories (units in meter). \textbf{Top}: Top-down view of true camera and object trajectories in evaluation sequences. The green trajectory represents the true object position in the Vicon reference frame. The red/blue trajectory segments represent the camera trajectory and if the object is static (blue) or dynamic (red) within the image. Black arrows point in the camera view direction. \textbf{Bottom}: True and estimated trajectories for our RigidFusion (with and without drift on motion priors), the baselines StaticFusion \cite{scona2018staticfusion} (with and without true motion priors) and Co-Fusion \cite{runz2017co}. Trajectories start at the origin (black solid dot) and end at the circle-cross marker. Our proposed method is closer to the ground truth trajectory even with drift on the motion priors (red, dashed), while StaticFusion fails even with true prior (blue, solid).}
    \label{fig:camera_trajectories}
\end{figure*}

We collected four sequences involving plane-parallel movement of the camera and the object within the camera frame. These sequences have different characteristics of camera and object motion (\Cref{tab:sequences}). \Cref{fig:camera_trajectories} (top) shows the 2D plane projection of the true trajectories.

\begin{table}
\centering
\begin{tabular}{l|cc|l}
\multicolumn{1}{c|}{\textbf{sequence}} & \multicolumn{2}{c|}{\textbf{frame motion}}       & \multicolumn{1}{c}{\textbf{difficulty}} \\
                                       & \textbf{camera} & \textbf{object}               &                                         \\ \hline
straight                            & straight        & orthogonal crossing           & low                                     \\
orbit                         & orbit        & rotation to camera            & medium                                  \\
overtake                          & straight         & rotation + parallel to camera & medium                                  \\
sideway                            & lateral         & orthogonal zig-zag crossing   & high                                   
\end{tabular}
\caption{Camera sequence description.}
\label{tab:sequences}
\end{table}

Our approach RigidFusion (RF) is compared against Joint-VO-SF (JF, \cite{jaimez2017fast}), StaticFusion (SF, \cite{scona2018staticfusion}), StaticFusion with true motion priors (SF true) and Co-Fusion (CF, \cite{runz2017co}). The quantitative evaluation in \Cref{tab:camera_eval} shows that our method outperforms the state-of-the-art on more difficult sequences. Although Co-Fusion achieves best results on the easier \textit{straight} sequence, it tends to over-segment dynamic objects and treats parts of the static background as dynamic. This effect is more dominant in the more difficult sequences, leading to worsen performance of CF.

\begin{table}[tb]
\centering
\begin{subtable}{\linewidth}
 \begin{tabular}{|c|c|ccccc|}
    \hline
    \multicolumn{1}{|c|}{RGB-D} & Motion prior & \multicolumn{5}{c|}{ Method} \\
\cline{3-7}    \multicolumn{1}{|c|}{sequence} & (drift) & \multicolumn{1}{c|}{JF} & \multicolumn{1}{c|}{SF} & \multicolumn{1}{c|}{SF true} & \multicolumn{1}{c|}{CF} & RF (ours) \\
    \hline
    straight & 17.6  & 48.4     & 34.8  & 14.5 & \textbf{3.84} & 7.57 \\
\hline    orbit & 44.2    & 52.0     & 87.7  & 19.9 & 14.2 & \textbf{5.74} \\
\hline    overtake & 8.93   & 59.6     & 52.6  & 23.6 & 23.0 & \textbf{5.39} \\
\hline   sideway & 51.1   & 55.3     & 70.1 & 38.1 & 48.2 & \textbf{13.1} \\
    \hline
    \end{tabular}%
    \caption{Trans. Absolute Trajectory Error RMSE (cm)}
  \label{tab:cam_ate}%
\end{subtable}

\vspace{0.1cm}

\begin{subtable}{\linewidth}
    \begin{tabular}{|c|c|ccccc|}
    \hline
    \multicolumn{1}{|c|}{RGB-D} & Motion prior & \multicolumn{5}{c|}{ Method} \\
\cline{3-7}    \multicolumn{1}{|c|}{sequence} & (drift) & \multicolumn{1}{c|}{JF} & \multicolumn{1}{c|}{SF} & \multicolumn{1}{c|}{SF true} & \multicolumn{1}{c|}{CF} & RF (ours) \\
    \hline
    straight & 6.02    & 18.5     & 24.3  & 12.9 & \textbf{5.54} & 6.05 \\
\hline    orbit & 6.03       & 13.4     & 25.2  & 5.78 & 8.47 & \textbf{5.1} \\
\hline    overtake & 6.34    & 19.1     & 27.4  & 11.3 & 18.9 & \textbf{4.78} \\
\hline    sideway & 6.01    & 21.7     & 42.3  & 9.87 & 17.0   & \textbf{8.03} \\
    \hline
    \end{tabular}%
    \caption{Trans. Relative Pose Error RMSE (cm/s)}
    \label{tab:cam_rpe}%
\end{subtable}
\setlength{\belowcaptionskip}{0.6cm}
\caption{ATE and RPE for camera-only sequences. \emph{Motion prior} represents the trajectory computed from prior motion with simulated drift to indicate the performance of simple kinematic odometry.
Our method with motion prior drift outperforms the state-of-the-art on difficult sequences, including SF with true motion prior (SF true), while Co-Fusion performs best on the easiest sequence.}
\label{tab:camera_eval}
\end{table}

\begin{figure*}[tb]
\centering
\setlength{\belowcaptionskip}{0cm}
\setlength{\tabcolsep}{0pt}
\newcommand{\h}{1.8cm}
\begin{tabular}{r|ccccc|c}
\textbf{}                                                                                           & \multicolumn{5}{c|}{\textbf{segmentation per frame ID}}                                                                                                                                                                                                                                                                                                                                                  & \textbf{reconstruction}                                       \\
                                                                                                    & 294                                                                          & 369                                                                          & 461                                                                          & 906                                                                          & 944                                                                          &                                                                      \\
\begin{tabular}[r]{@{}r@{}}RGB input \\(dynamic segment\\ marked in red)\end{tabular}\hspace{0.5cm} & \includegraphics[height=\h,valign=m]{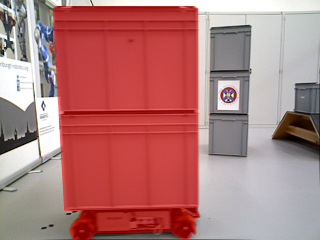}       & \includegraphics[height=\h,valign=m]{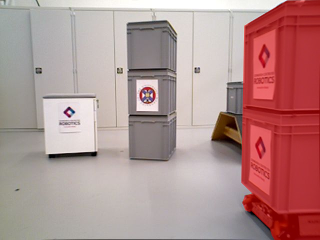}       & \includegraphics[height=\h,valign=m]{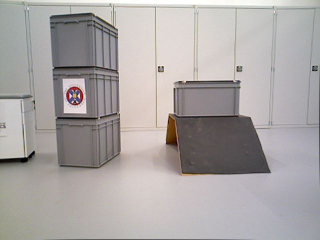}        & \includegraphics[height=\h,valign=m]{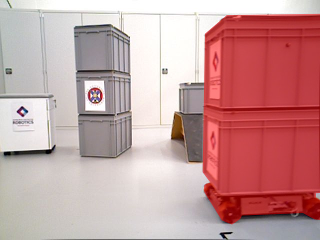}       & \includegraphics[height=\h,valign=m]{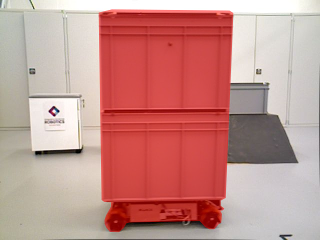}       & \includegraphics[height=\h,valign=m]{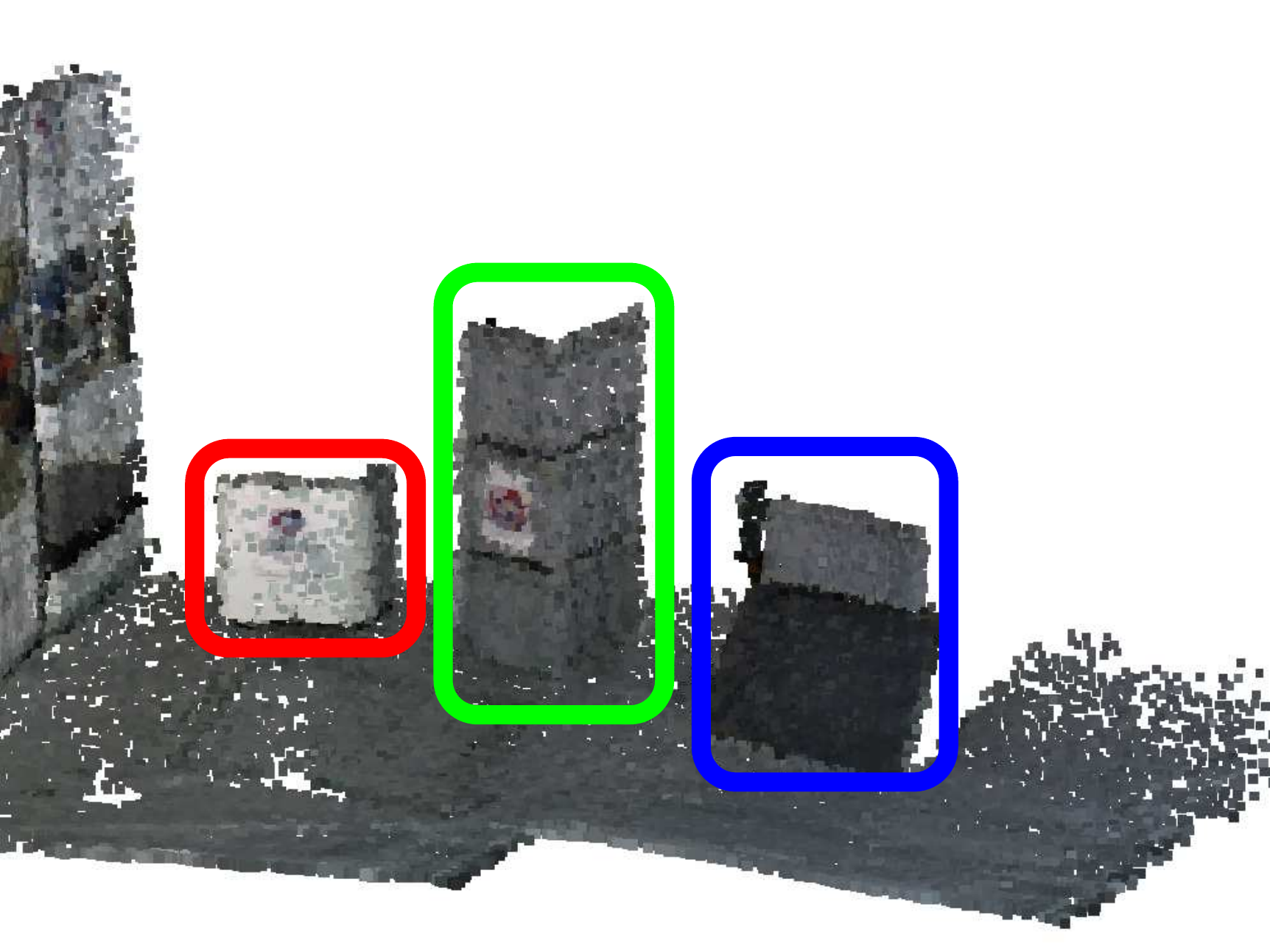} \\
CF\hspace{0.5cm}                                                                                    & \includegraphics[height=\h,valign=m]{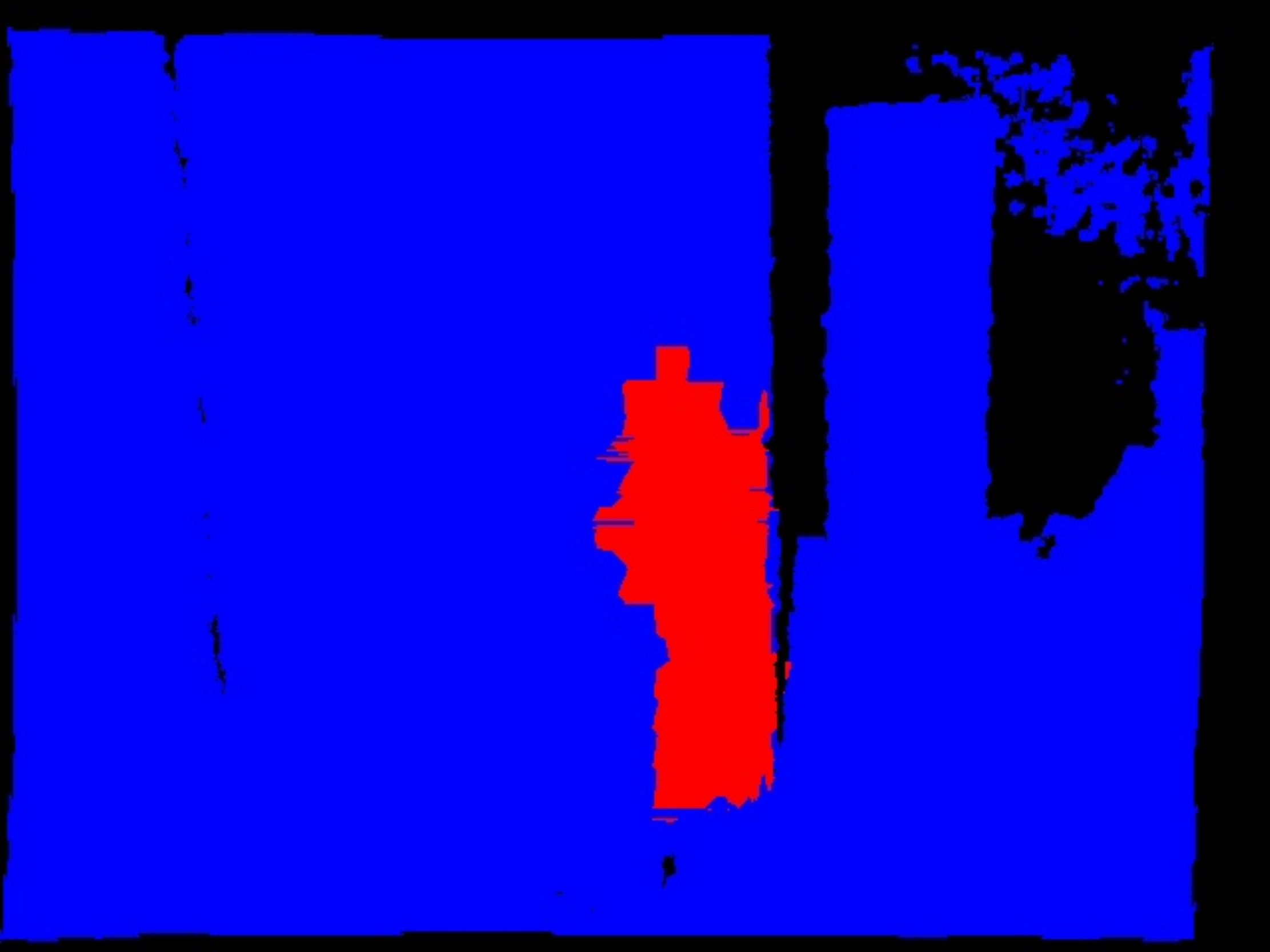}         & \includegraphics[height=\h,valign=m]{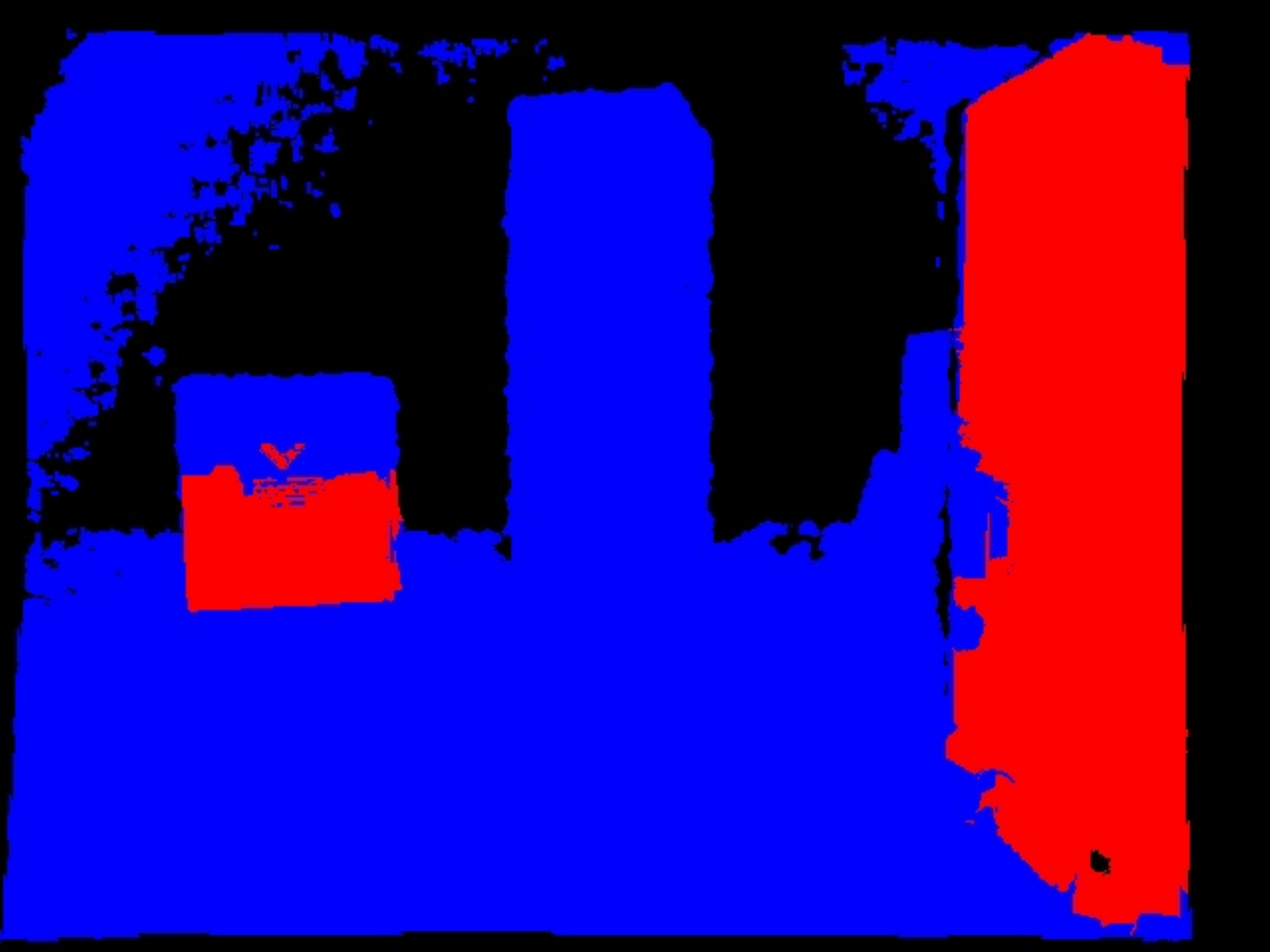}         & \includegraphics[height=\h,valign=m]{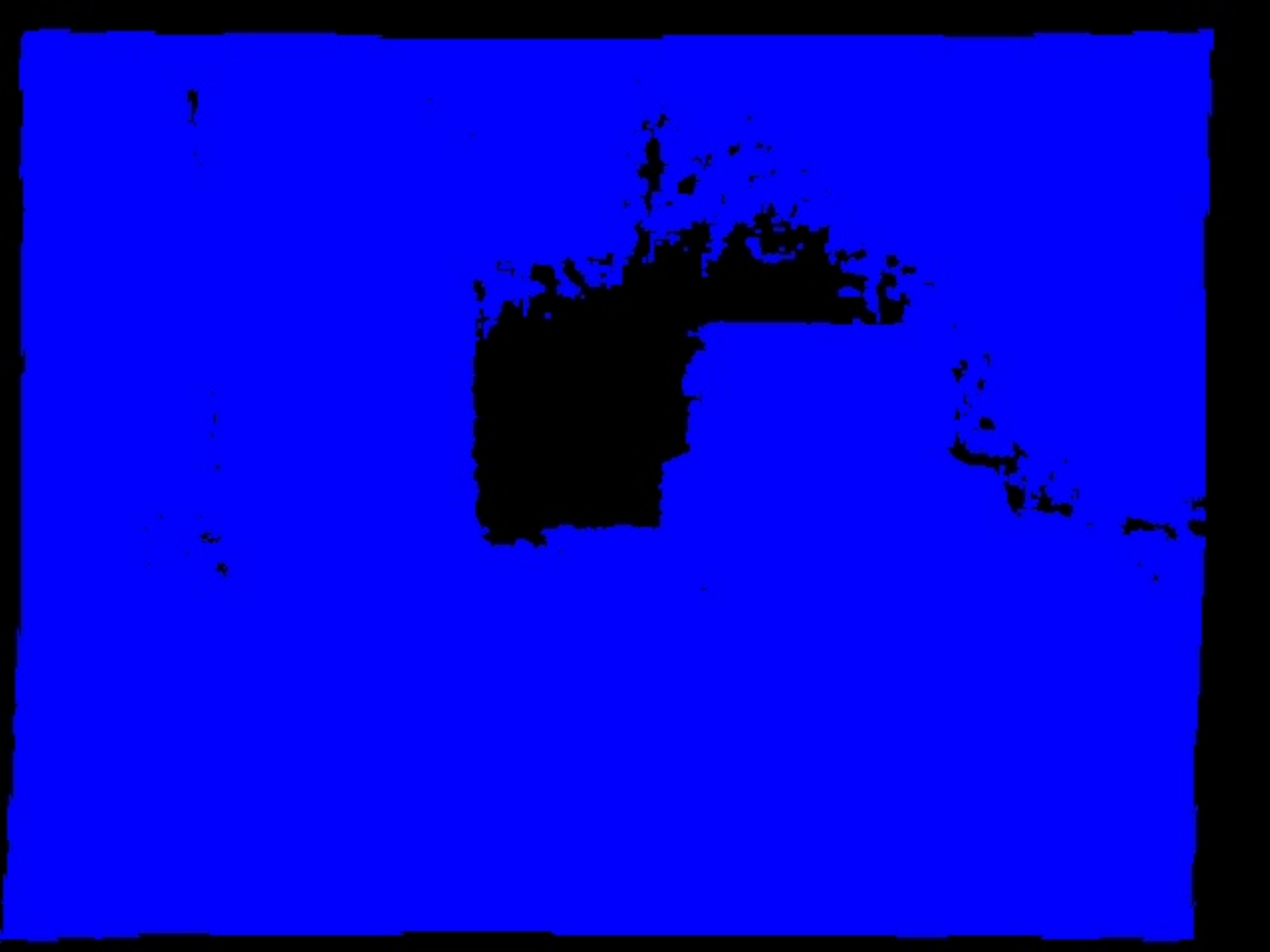}         & \includegraphics[height=\h,valign=m]{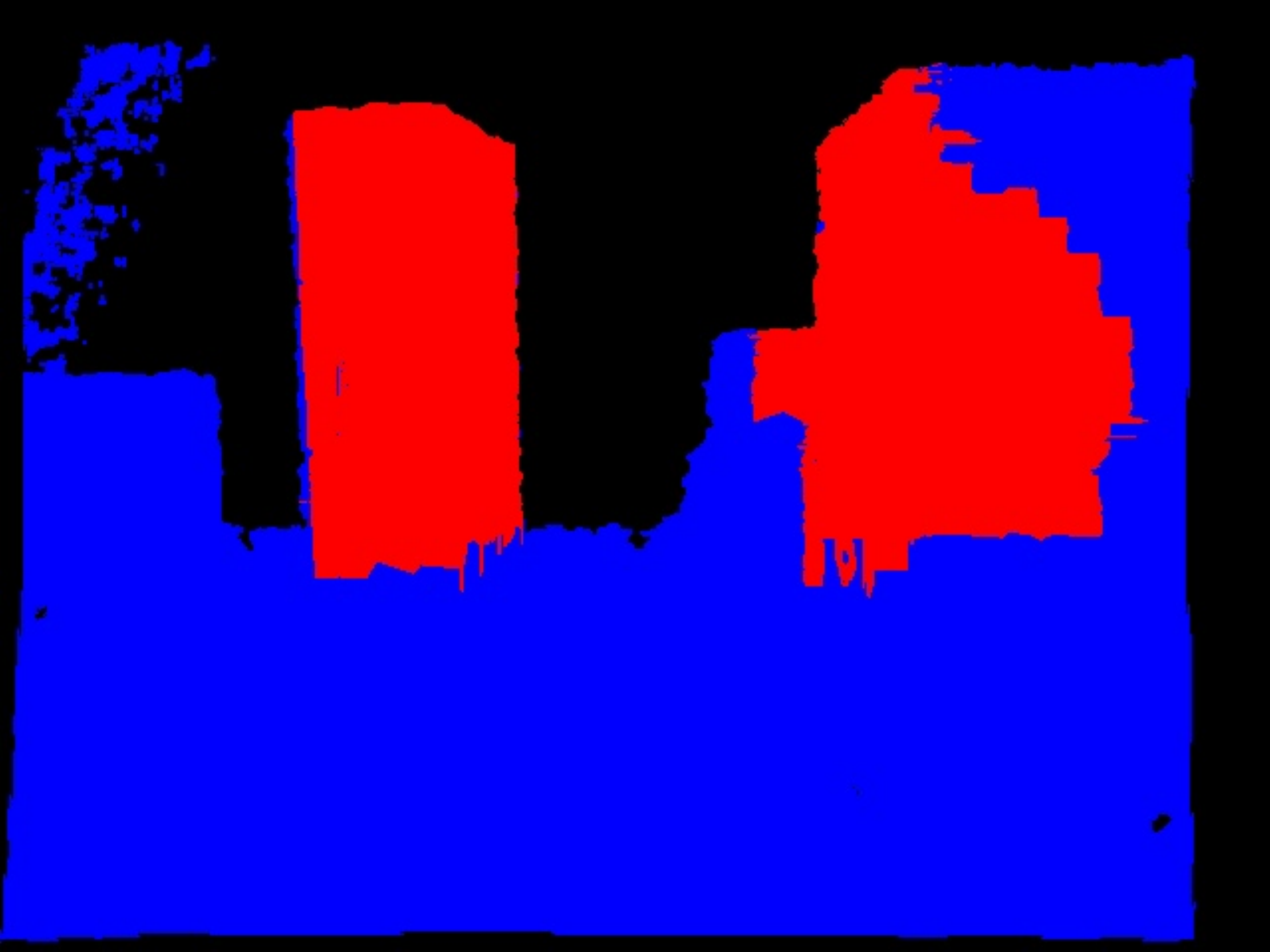}         & \includegraphics[height=\h,valign=m]{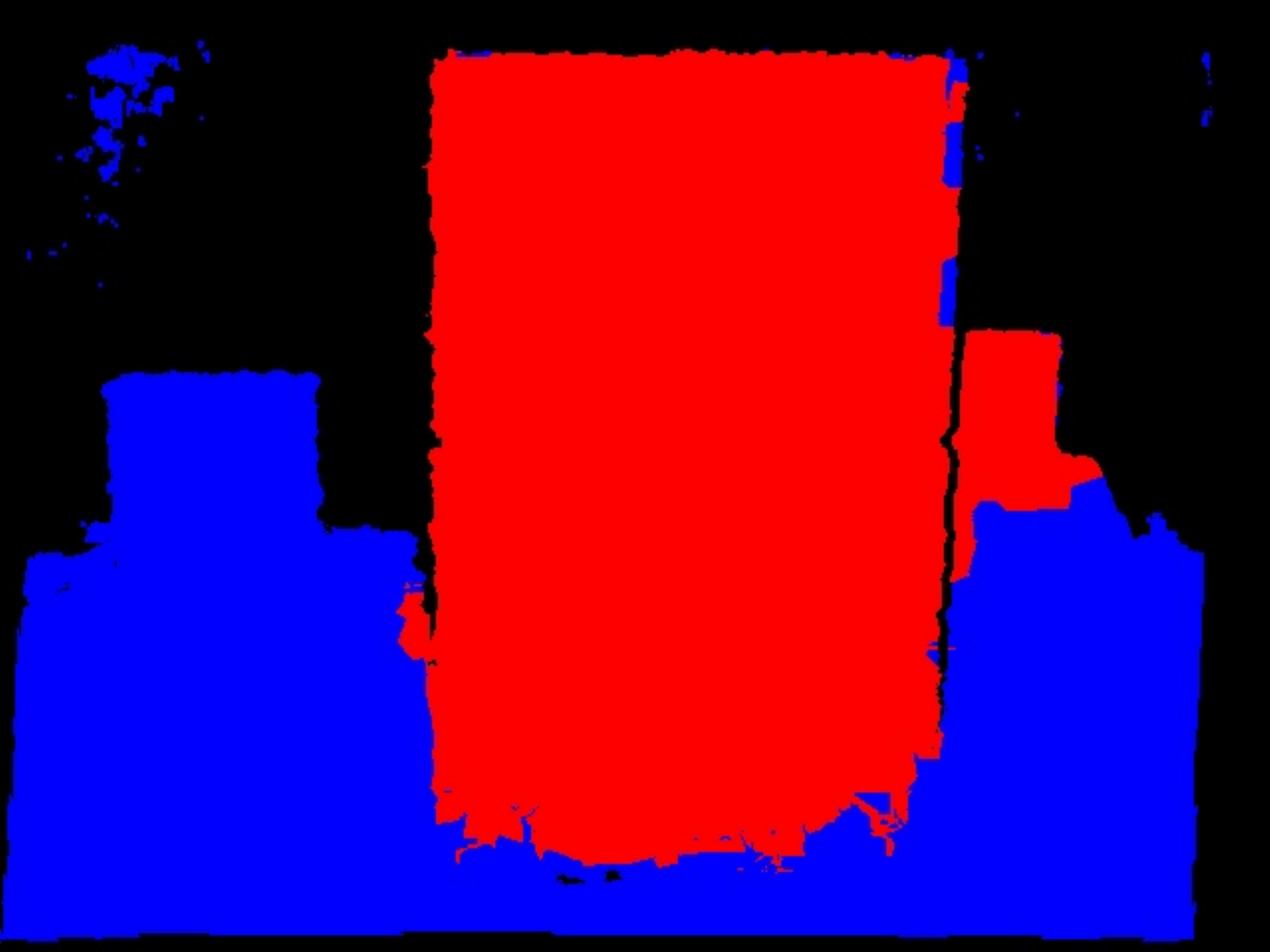}         & \includegraphics[height=\h,valign=m]{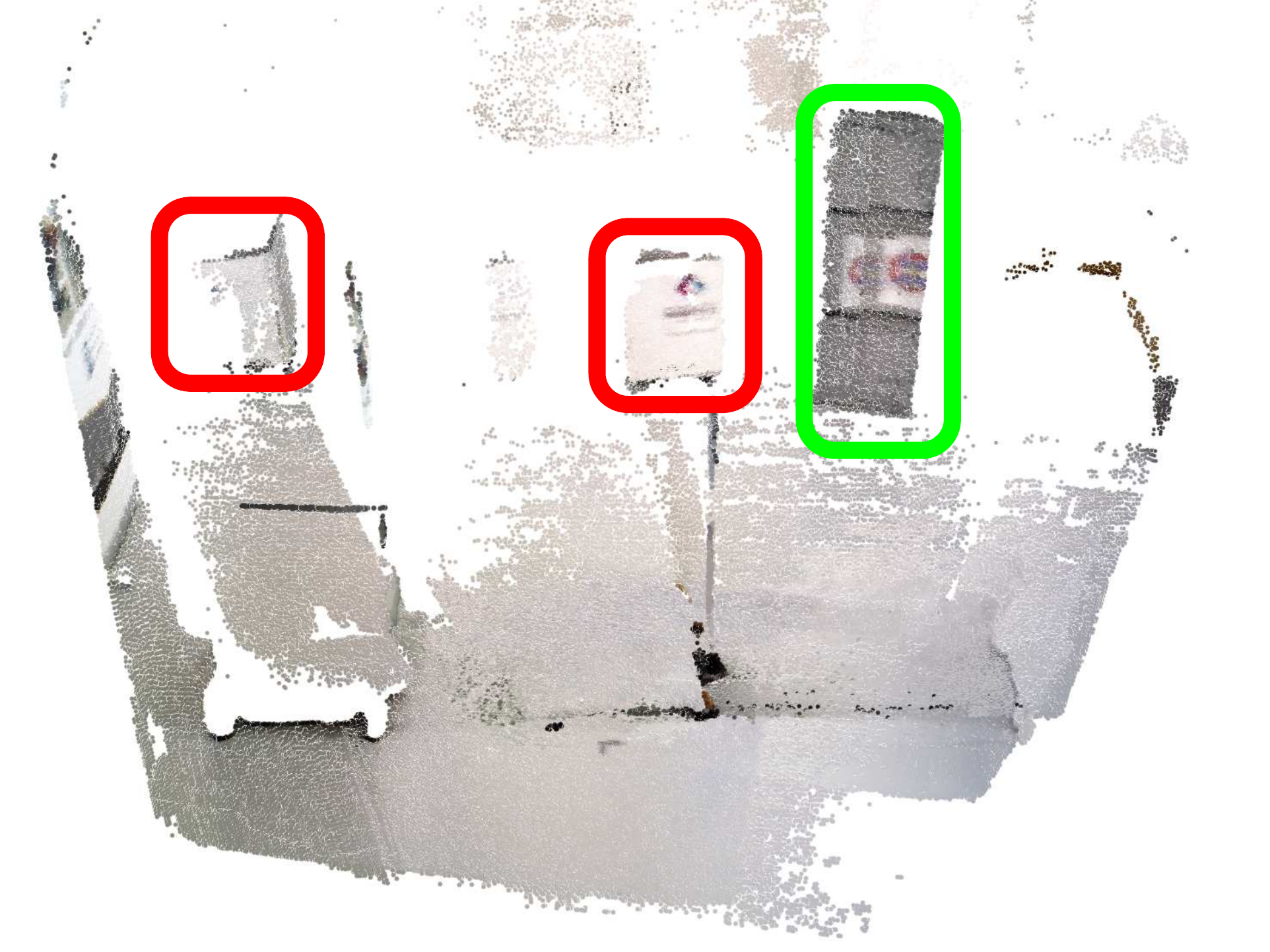}  \\

\begin{tabular}[r]{@{}r@{}}SF \\(baseline)\end{tabular}\hspace{0.5cm}                                & \includegraphics[height=\h,valign=m]{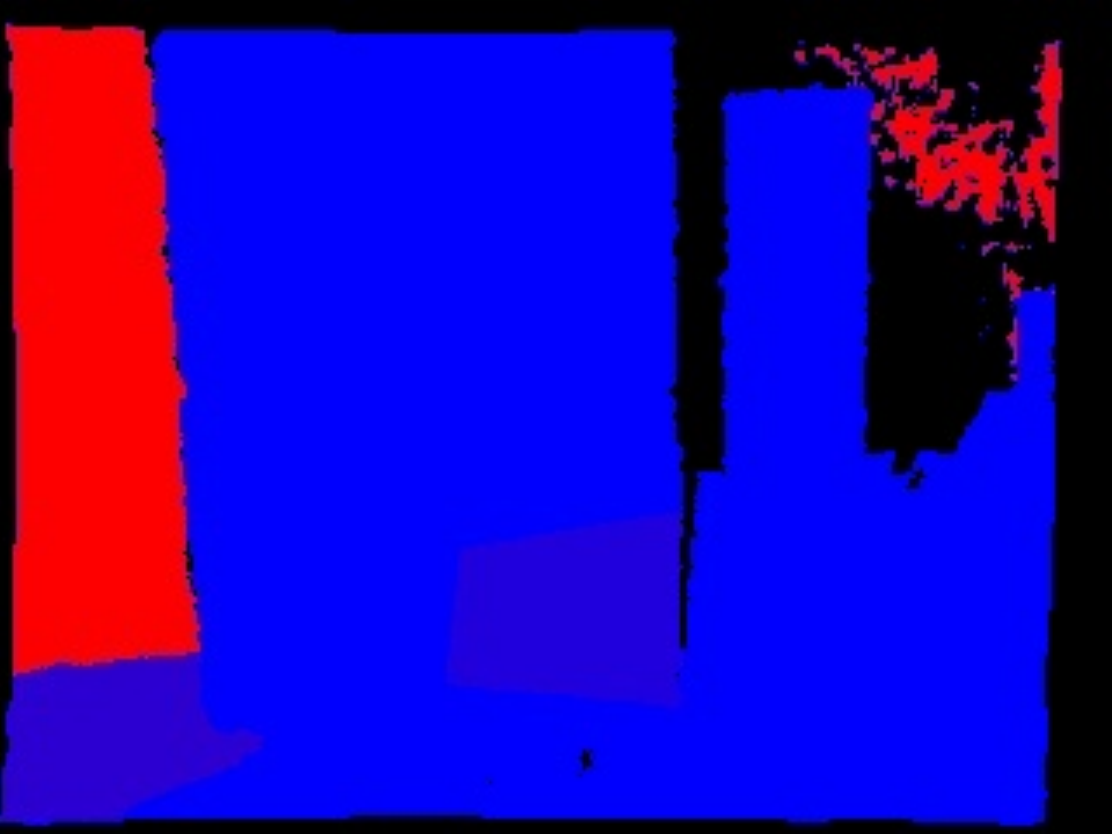} & \includegraphics[height=\h,valign=m]{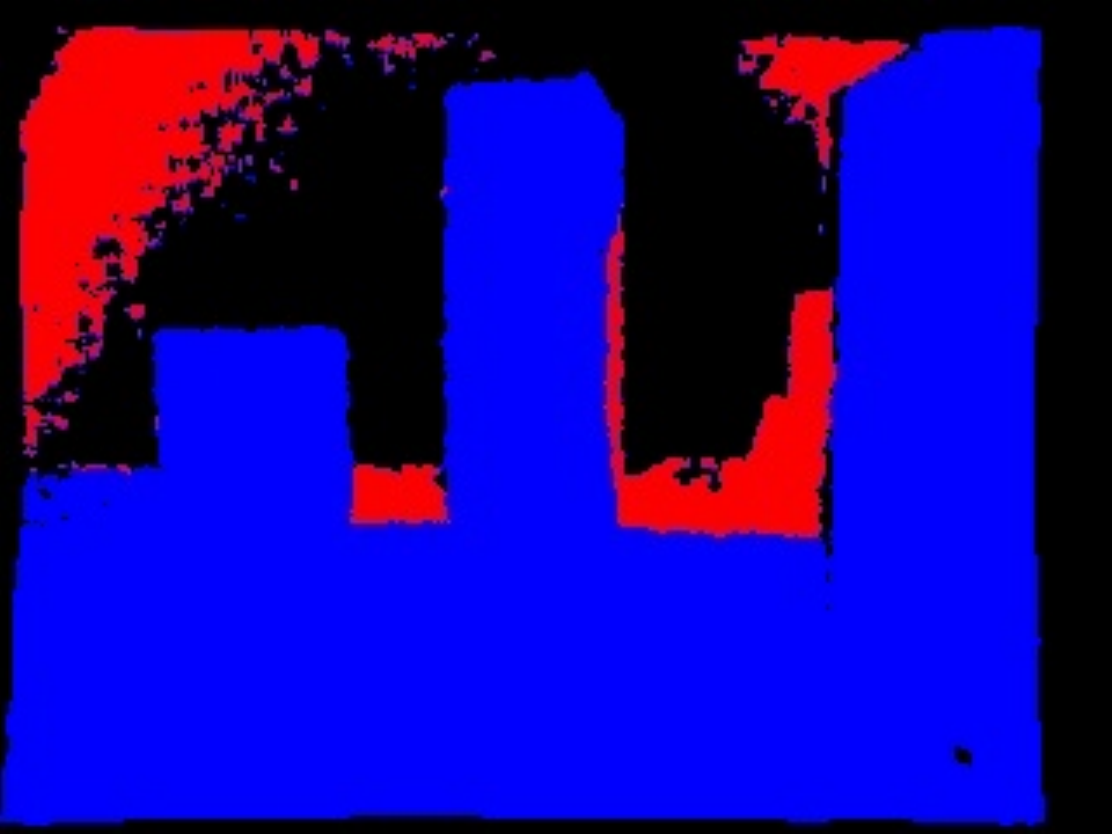} & \includegraphics[height=\h,valign=m]{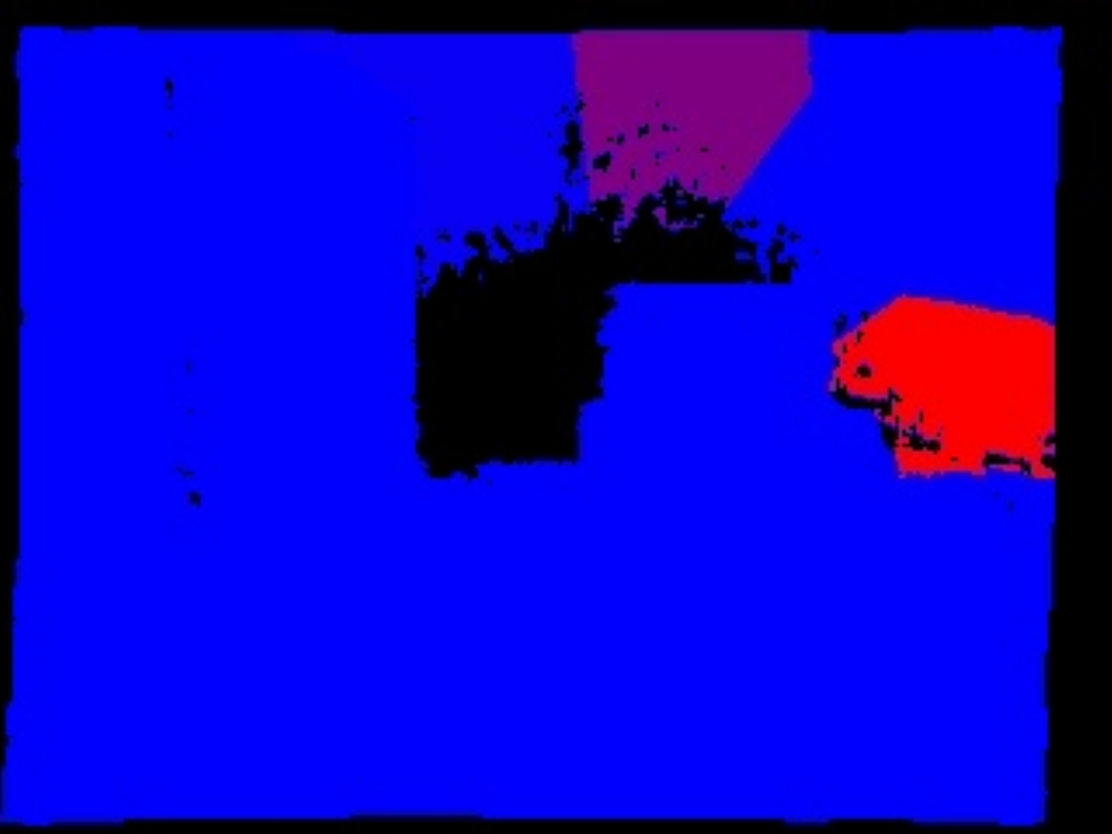} & \includegraphics[height=\h,valign=m]{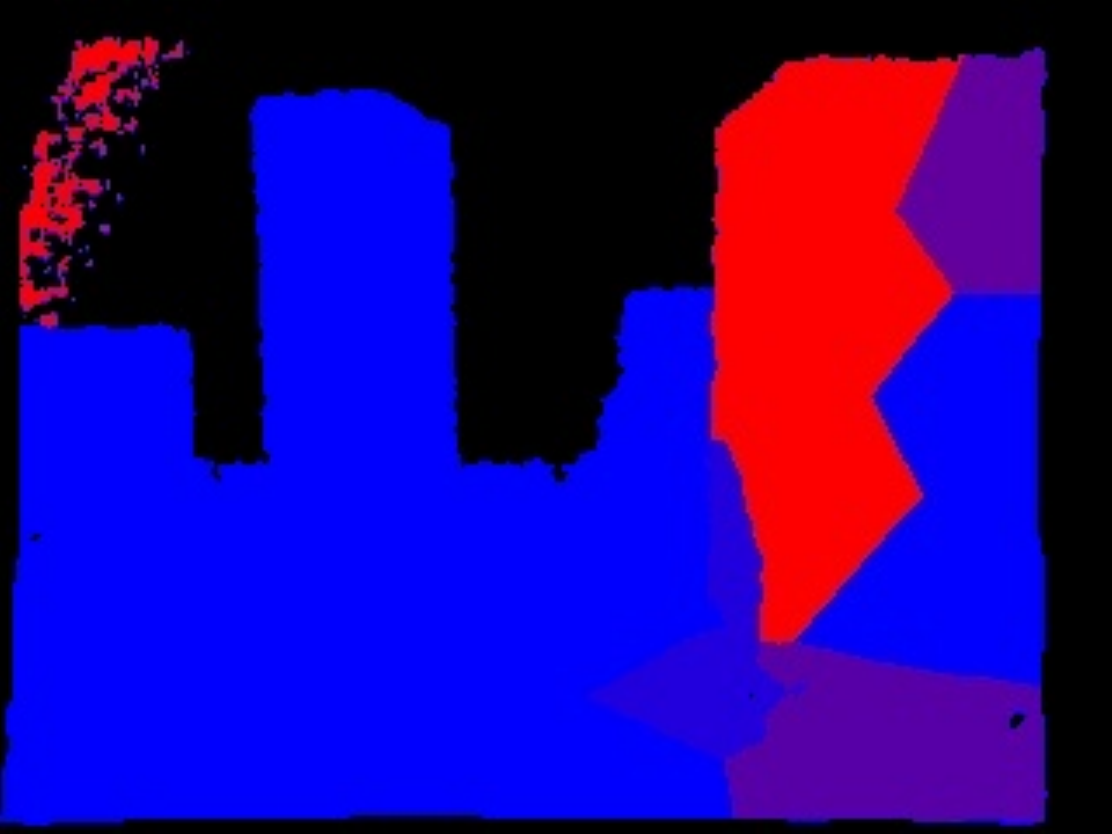} & \includegraphics[height=\h,valign=m]{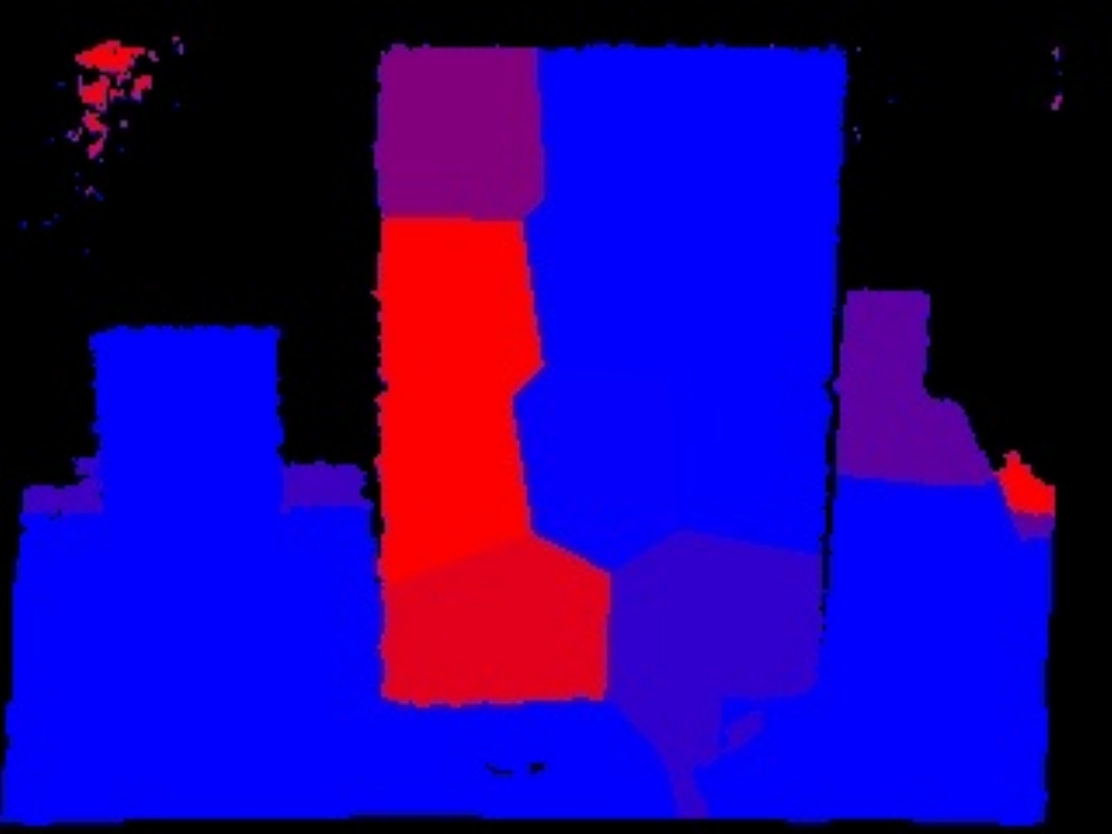} & \includegraphics[height=\h,valign=m]{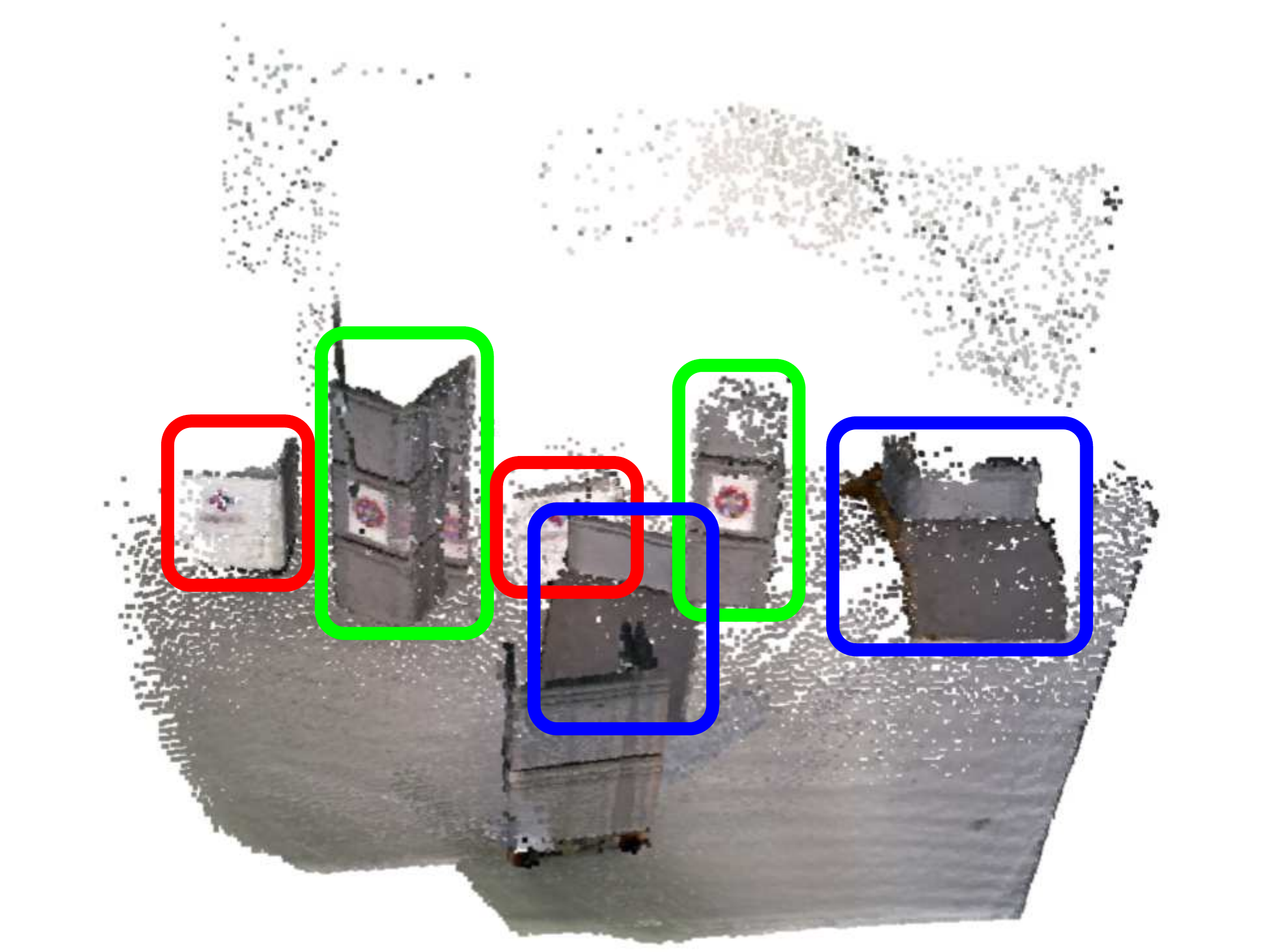}  \\
\begin{tabular}[r]{@{}r@{}}SF \\(true motion prior)\end{tabular}\hspace{0.5cm}                              & \includegraphics[height=\h,valign=m]{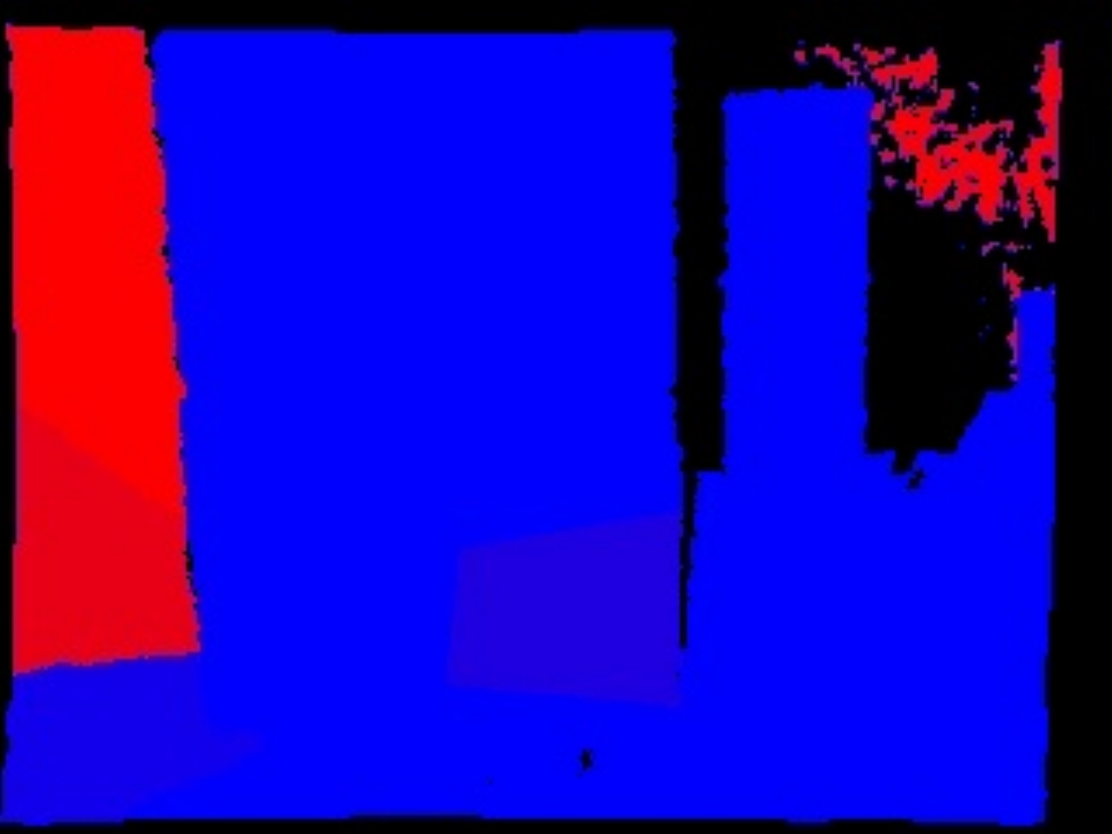}     & \includegraphics[height=\h,valign=m]{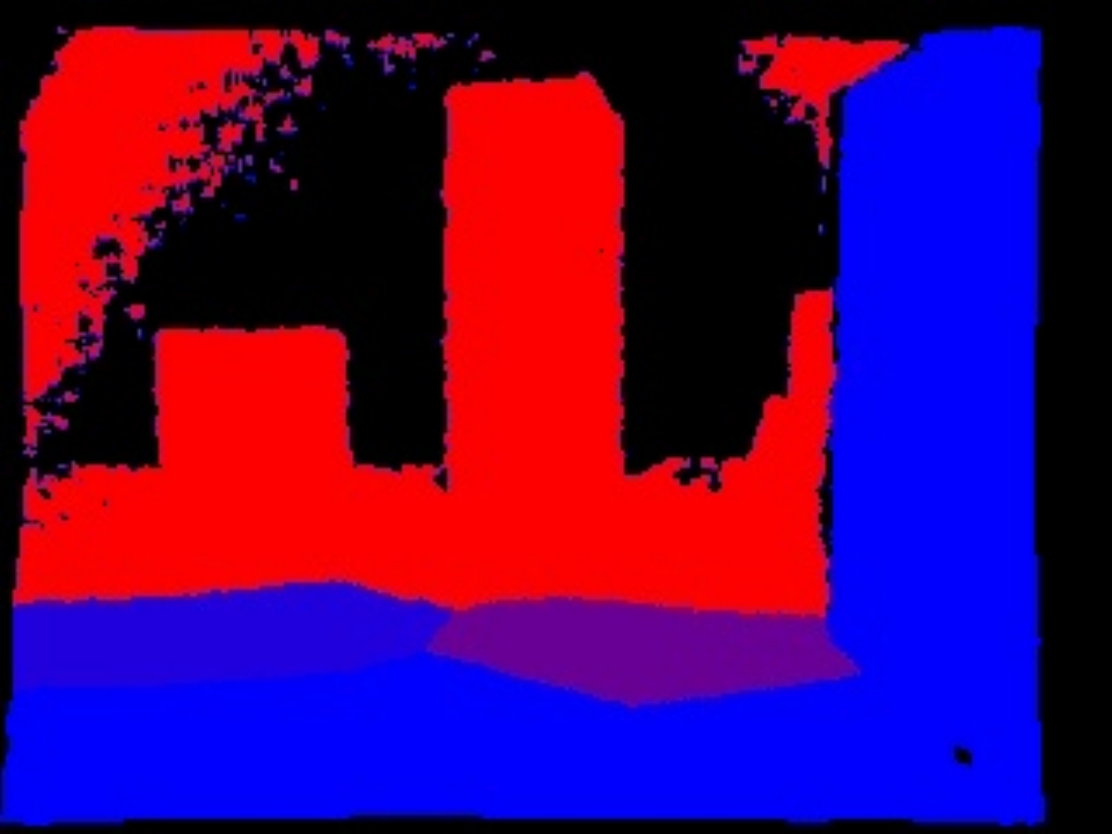}     & \includegraphics[height=\h,valign=m]{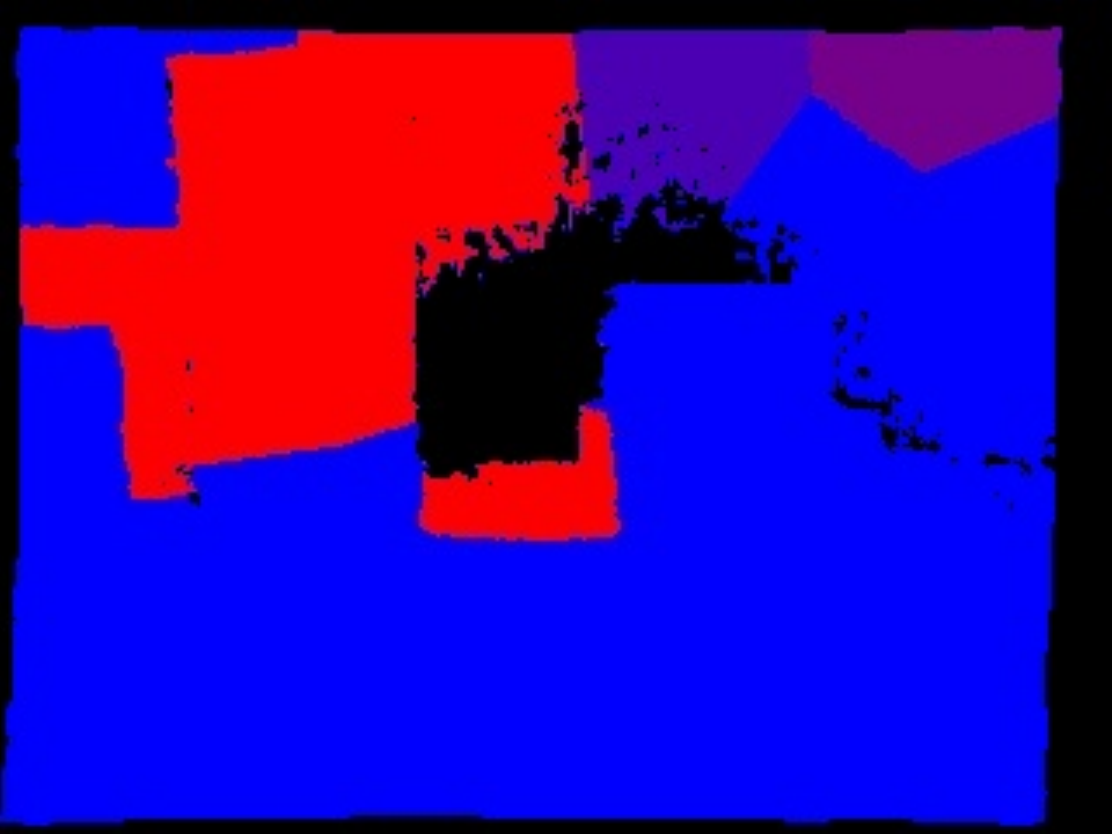}     & \includegraphics[height=\h,valign=m]{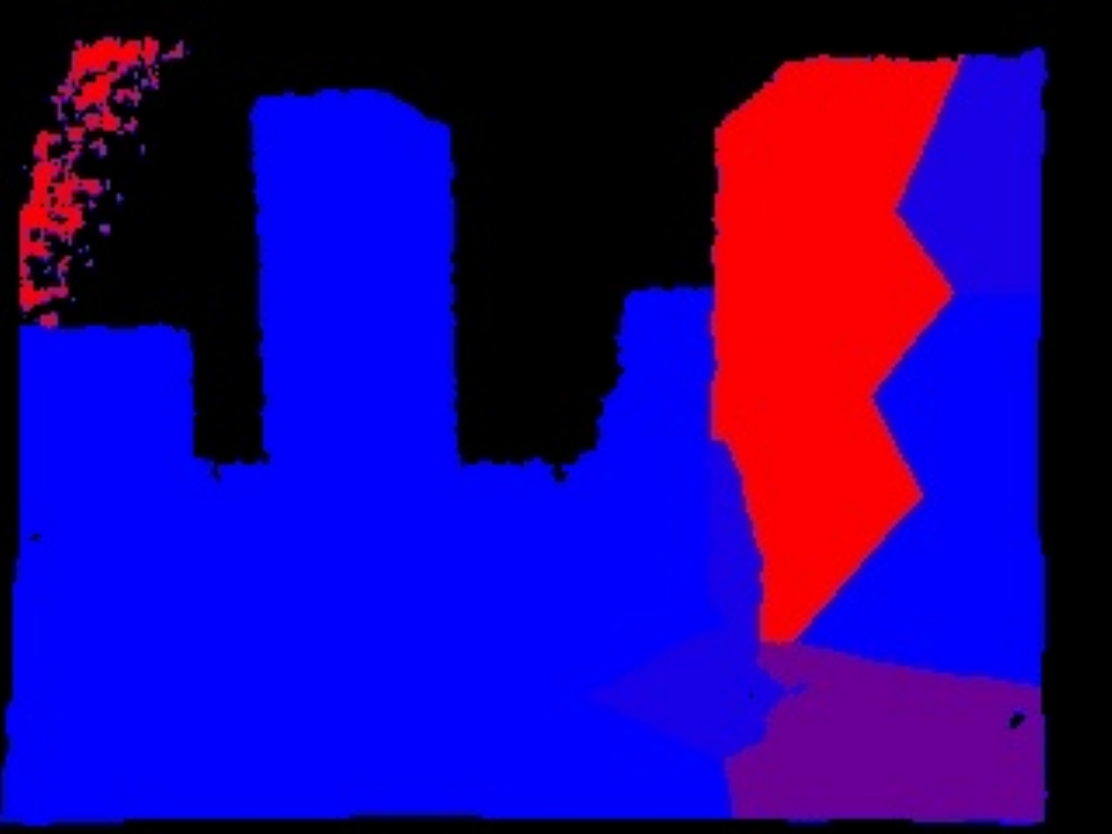}     & \includegraphics[height=\h,valign=m]{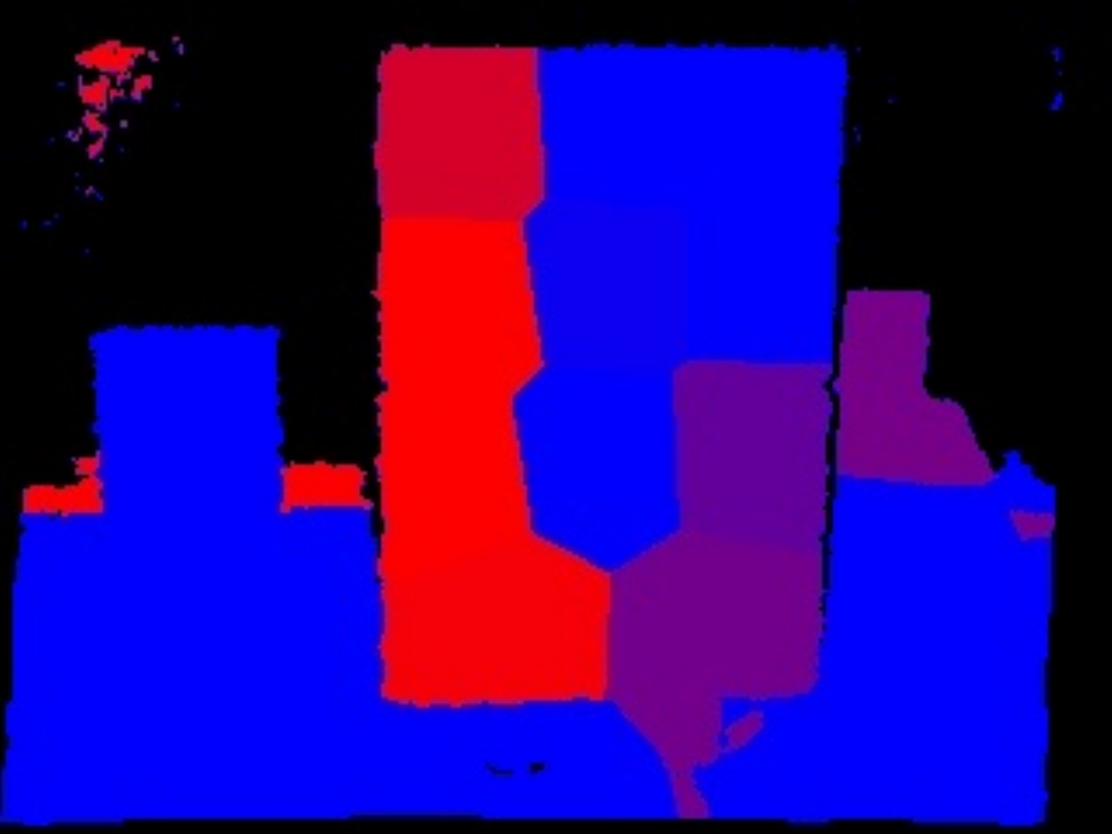}     & \includegraphics[height=\h,valign=m]{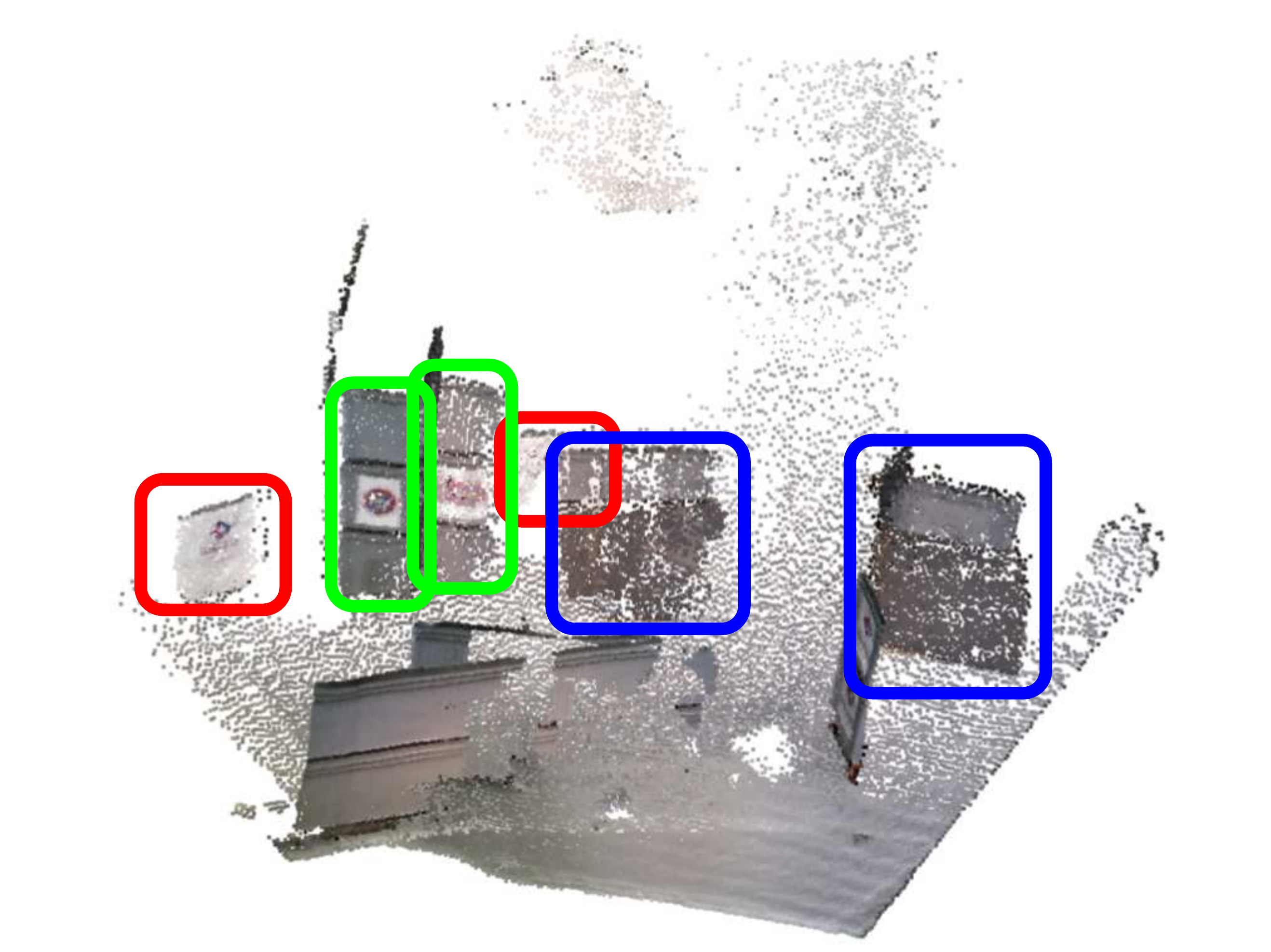}      \\
\begin{tabular}[r]{@{}r@{}}RF (ours)\\(motion prior drift)\end{tabular}\hspace{0.5cm}                      & \includegraphics[height=\h,valign=m]{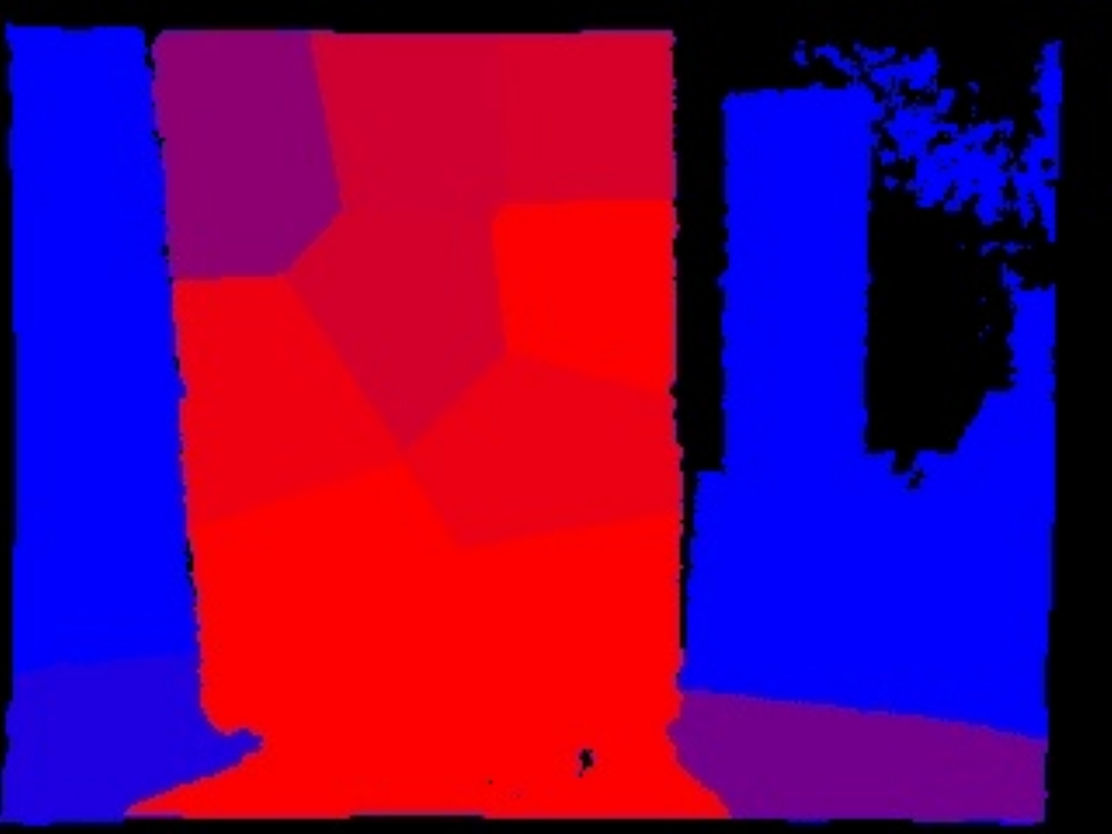}    & \includegraphics[height=\h,valign=m]{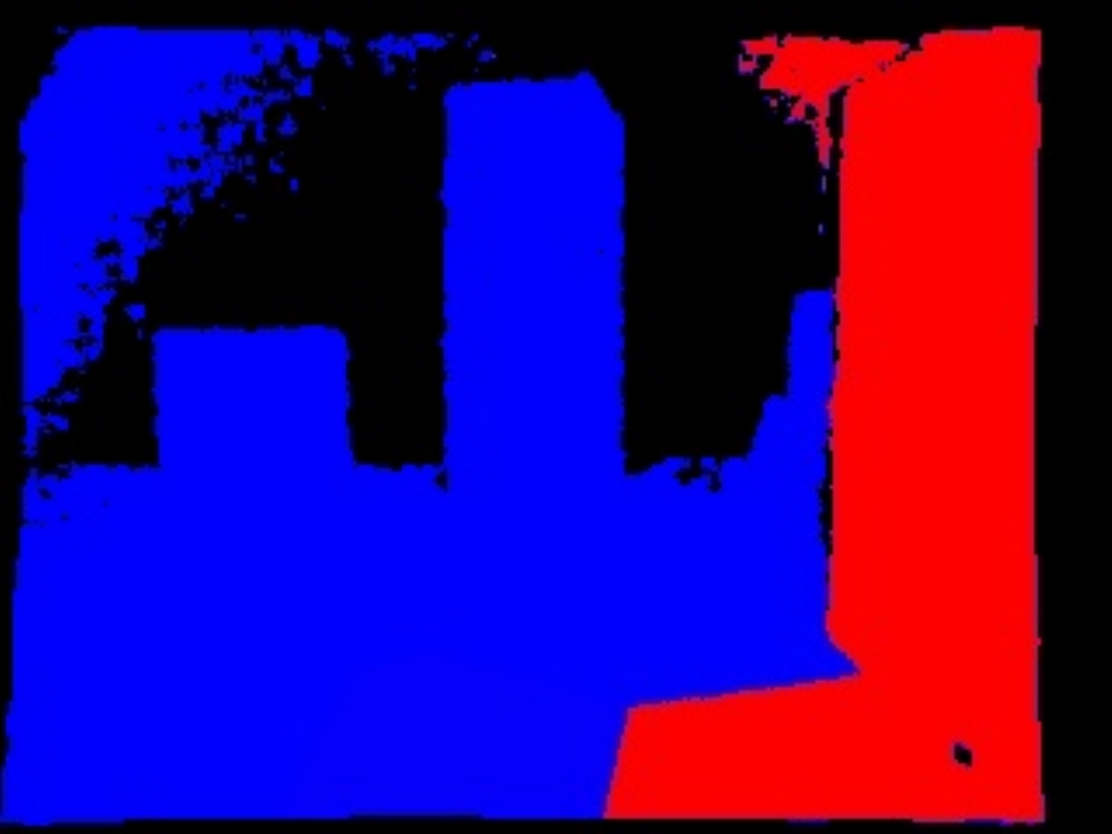}    & \includegraphics[height=\h,valign=m]{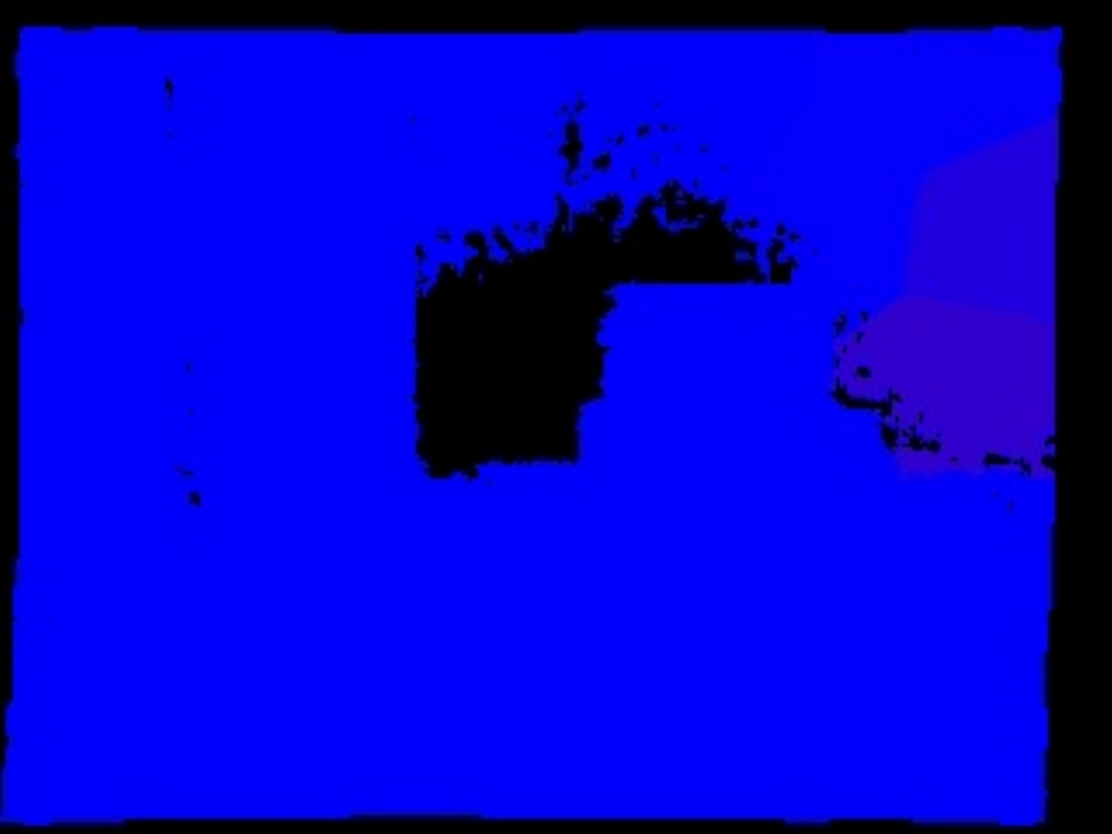}    & \includegraphics[height=\h,valign=m]{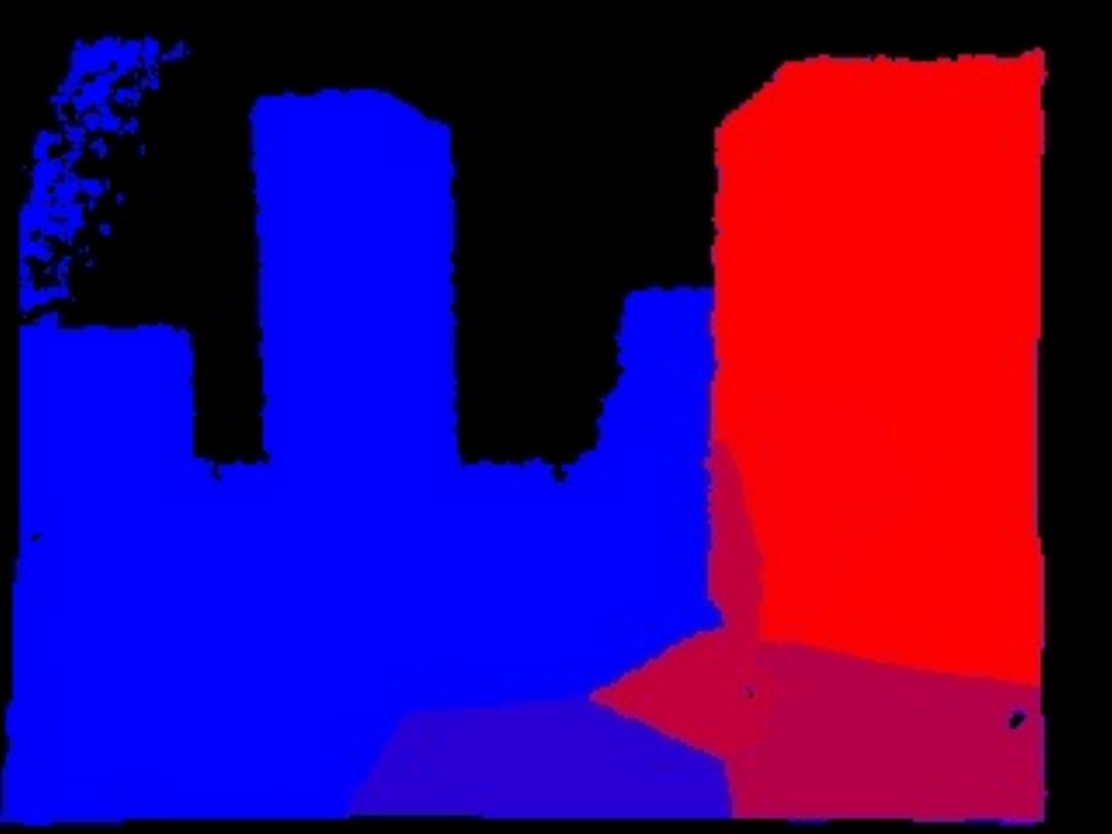}    & \includegraphics[height=\h,valign=m]{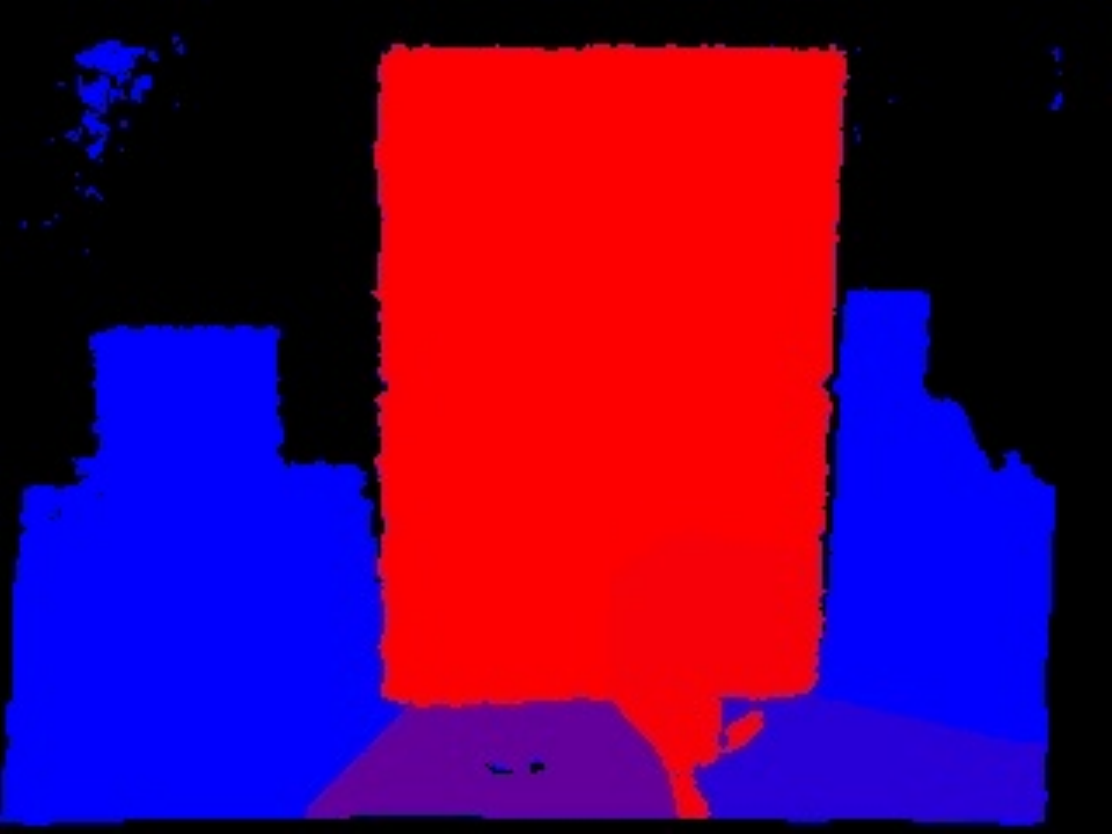}    & \includegraphics[height=\h,valign=m]{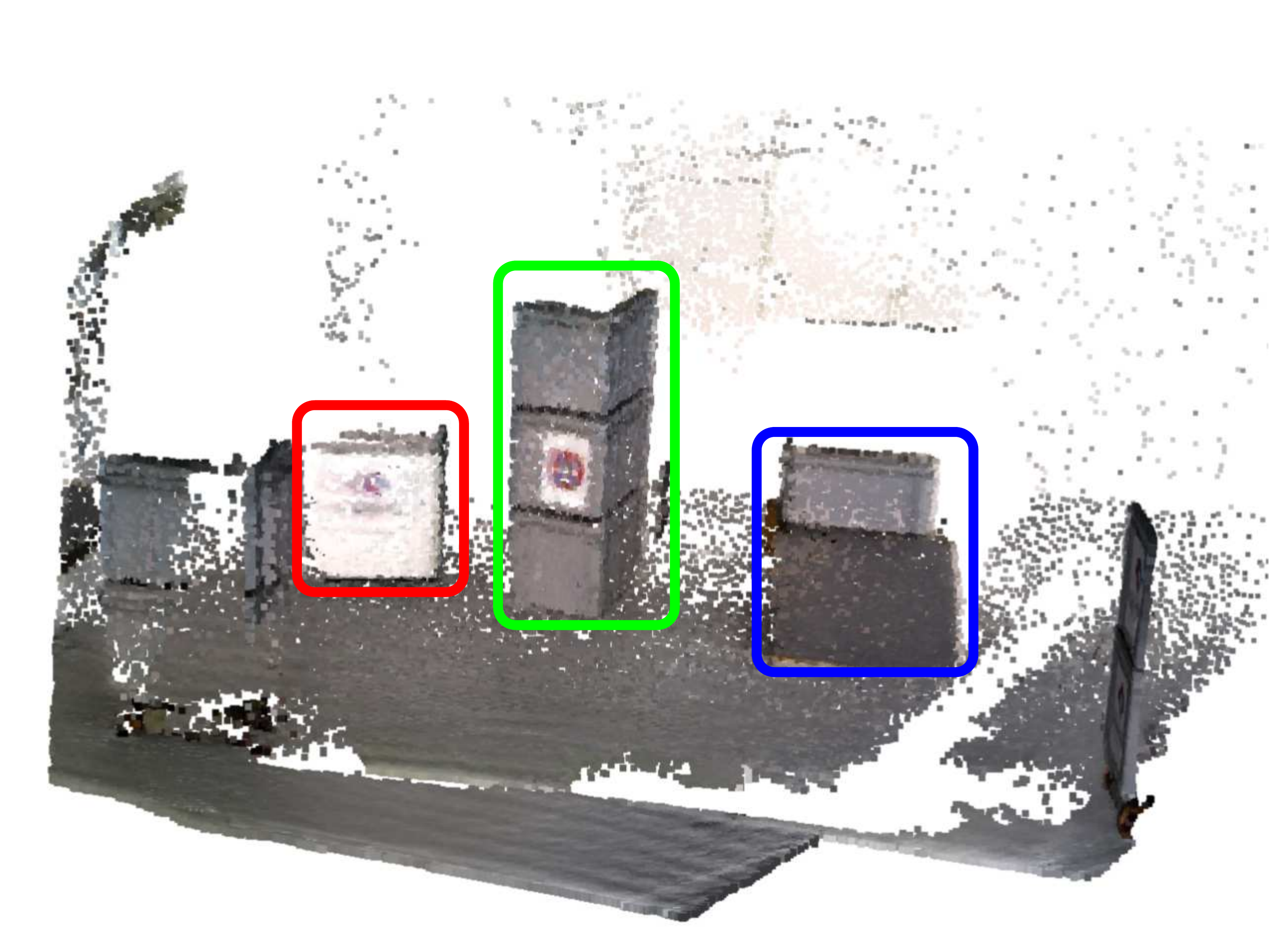}    
\end{tabular}
\caption{Segmentation and 3D reconstructed background for our proposed algorithm RigidFusion (RF), StaticFusion (SF) \cite{scona2018staticfusion} (with and without true motion priors), Joint-VO-SF (JF) \cite{jaimez2017fast} and Co-Fusion (CF) \cite{runz2017co} on camera-only sequence \emph{sideway}. Our proposed method is the only one that can consistently segment the large rigid dynamic object (compare first row with highlighted boxes against red dynamic segmentation) and reconstruct the background even the motion priors have a significant drift.}
\label{fig:segmentation}
\end{figure*}

The visualisation of the estimated trajectories in \Cref{fig:camera_trajectories} (bottom) confirms that our method outperforms the state-of-the-art in dynamic scenes. The improved localisation performance stems from a better segmentation of dynamic parts in the image (\Cref{fig:segmentation}). In our frame-to-frame odometry setting, the improved motion segmentation performance directly affects the estimation performance and additionally leads to better a reconstruction of the static environment.

\subsection{Robot Experiments}

\begin{figure}[tb]
    \centering
    \includegraphics[width=\linewidth]{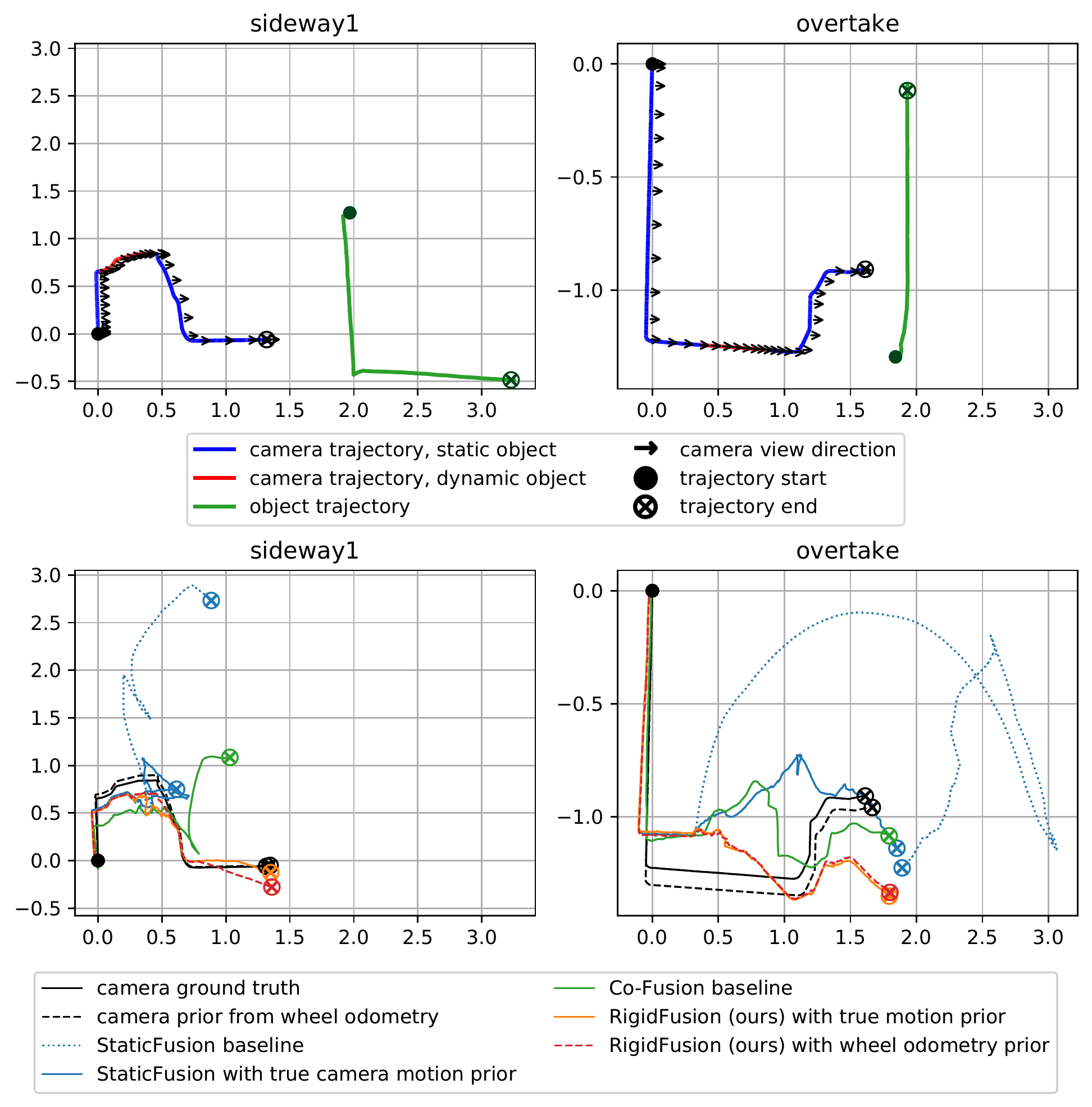}
    \caption{True (top) and estimated (bottom) trajectories (units in meter). Our method (RF) outperforms all state-of-the art methods. Although CF has closer end-position in x-y plane, it has a larger drift in the z position than RF.}
    \label{fig:ada_trajectories}
\end{figure}

In four additional robotic experiments, we use the camera on the floating base of an omnidirectional robot and replace the simulated drift with wheel odometry. The true trajectories of two of these sequences are shown in \Cref{fig:ada_trajectories} (top).

The quantitative results in \Cref{tab:ada_eval} show that using real wheel odometry as motion priors, RF outperforms all other four methods in terms of both ATE and RPE on all four sequences.
The estimated trajectories for sequences \emph{sideway1} and \emph{overtake} are shown in \Cref{fig:ada_trajectories} (bottom).

\begin{table}[tb]
\centering
\begin{subtable}{\linewidth}
 \begin{tabular}{|c|c|ccccc|}
    \hline
    \multicolumn{1}{|c|}{RGB-D} & Wheel & \multicolumn{5}{c|}{ Method} \\
\cline{3-7}    \multicolumn{1}{|c|}{sequence} & odometry & \multicolumn{1}{c|}{JF} & \multicolumn{1}{c|}{SF} & \multicolumn{1}{c|}{SF true} & \multicolumn{1}{c|}{CF} & RF (ours) \\
    \hline
sideway1 &    2.27    & 37.7  & 62.8  & 36.8 & 32.1 & \textbf{7.58} \\
\hline overtake &    3.16   & 23.8  & 79.1  & 24.7 & 16.5  & \textbf{14.0} \\
\hline straight &    3.64   & 51.9  & 86.3  & 21.9 & 12.3 & \textbf{7.98} \\
\hline sideway2 &    3.21   & 53.8  & 54.2  & 34.1 & 32.9 & \textbf{10.7} \\
    \hline
    \end{tabular}%
    \caption{Trans. Absolute Trajectory Error RMSE (cm)}
  \label{tab:ada_ate}%
\end{subtable}

\vspace{0.1cm}

\begin{subtable}{\linewidth}
    \begin{tabular}{|c|c|ccccc|}
    \hline
      RGB-D    &  Wheel     & \multicolumn{5}{c|}{Method} \\
\cline{3-7}   sequence & odometry & \multicolumn{1}{c|}{JF} & \multicolumn{1}{c|}{SF} & \multicolumn{1}{c|}{SF true} & \multicolumn{1}{c|}{CF} & RF (ours) \\
    \hline
sideway1 &   0.77    & 19.4  & 34.7  & 15.9 & 18.6 & \textbf{3.66} \\
\hline overtake &    0.74    & 18.7  & 41.7  & 10.4 & 6.78 & \textbf{2.06} \\
\hline straight &    1.13  & 39.8  & 84.2  & 13.7  & 10.8  & \textbf{8.67} \\
\hline sideway2 &    1.14    & 22.2  & 57.5  & 18.2 & 12.9  & \textbf{6.68} \\
    \hline
    \end{tabular}%
    \caption{Trans. Relative Pose Error RMSE (cm/s)}
    \label{tab:ada_rpe}%
\end{subtable}
\caption{ATE and RPE for sequences collected with Ada. The camera motion prior is estimated from the wheel odometry. Our method (RF) outperforms all compared dynamic SLAM methods when using real wheel odometry.}
\label{tab:ada_eval}
\end{table}

\subsection{Object Reconstruction}
\begin{figure}
    \centering
    \setlength{\tabcolsep}{0pt}
    \begin{tabular}{cccc}
    & camera/orbit & camera/overtake& robot/sideway2 \\
    \hline
    \rotatebox[origin=c]{90}{CF}\hspace{0.1cm} & \includegraphics[width=0.3\linewidth,valign=m]{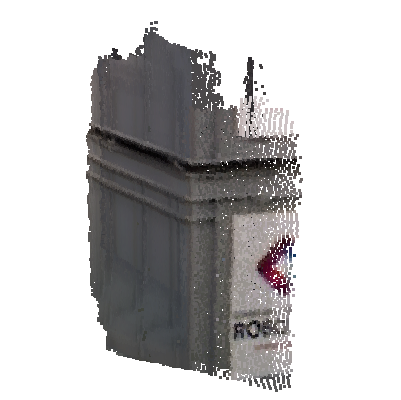}        & \includegraphics[width=0.3\linewidth,valign=m]{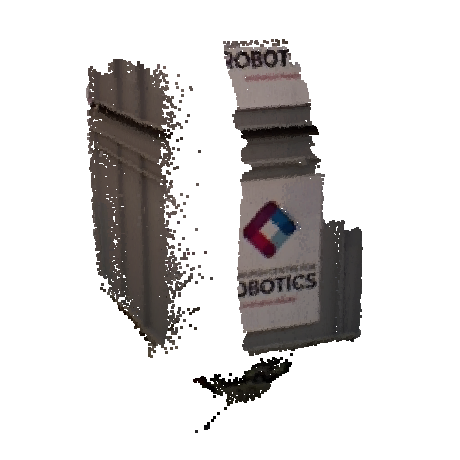}        & \includegraphics[width=0.3\linewidth,valign=m]{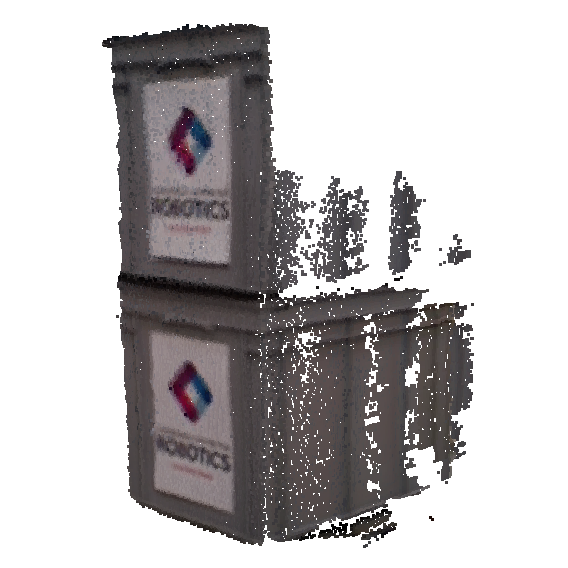}        \\
    \hline
    \rotatebox[origin=c]{90}{RF (ours)}\hspace{0.1cm}  & \includegraphics[width=0.3\linewidth,valign=m]{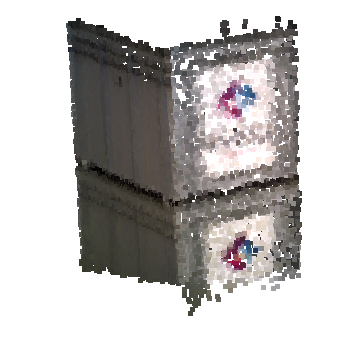} & \includegraphics[width=0.3\linewidth,valign=m]{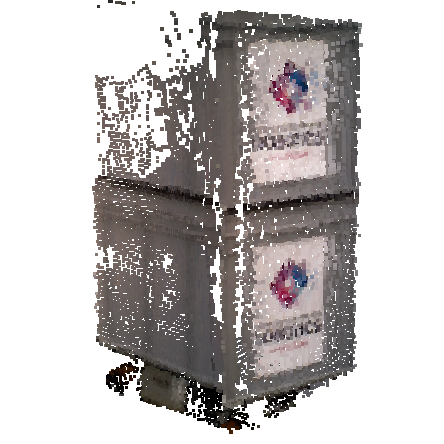} & \includegraphics[width=0.3\linewidth,valign=m]{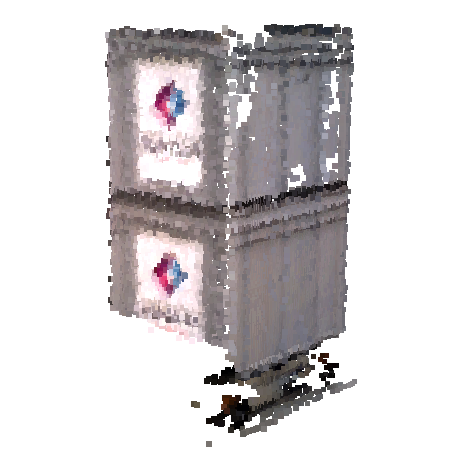} \\
    \hline
    \end{tabular}
    \caption{Reconstructed dynamic object. CF can only reconstruct parts of the dynamic object, while RF reconstructs a more complete model with inaccurate wheel odometry.}
    \label{fig:ada_dynamic_reconstruction}
\end{figure}

We compare the reconstructed dynamic object for CF and RF in \Cref{fig:ada_dynamic_reconstruction}. Since CF tends to over-segment objects, we only show the first detected model. Results show that RF generates a more complete dynamic model than CF. This suggests that the segmentation estimated by RF is consistent over time and more accurate than CF.

\subsection{Impact of Odometry Drift on Trajectory Estimation}
\begin{figure}
    \centering
    \includegraphics[width=\linewidth]{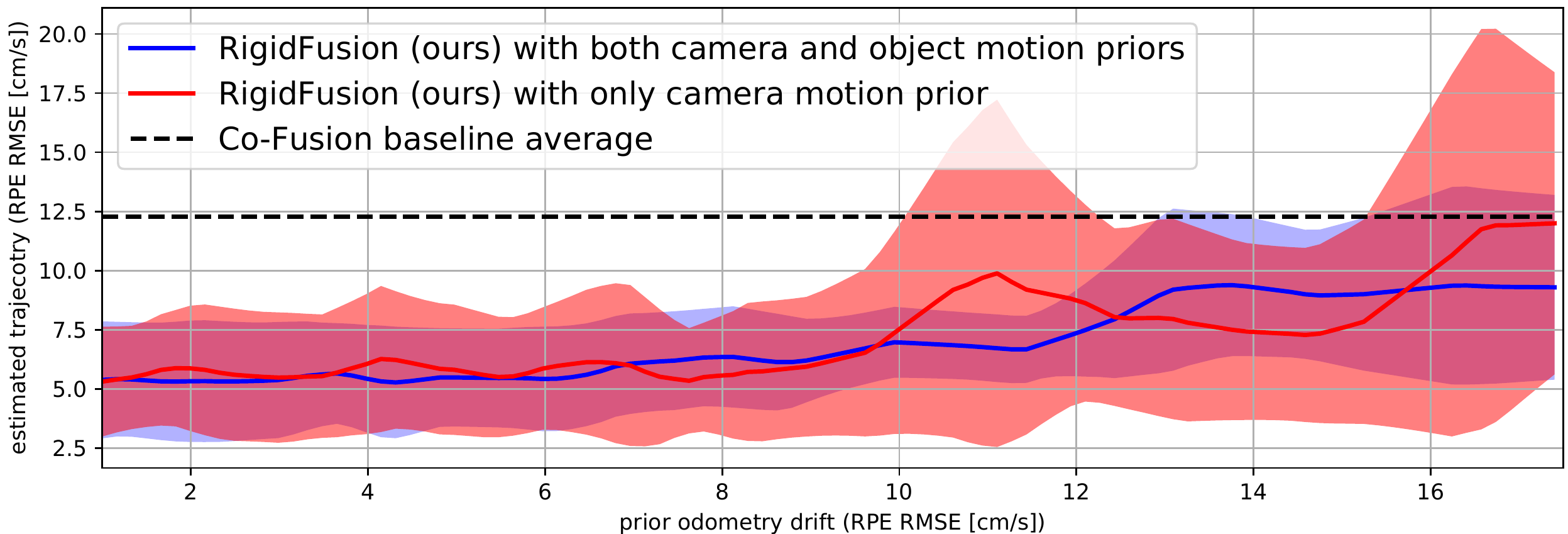}
    \caption{RPE of the estimated trajectories impacted by the drift magnitude of wheel odometry. Our method can handle up to 17 cm/s drift without the object motion prior (solid red) before breaking down to comparable results with CF. Using both motion priors (solid blue), RF has a even better performance and stronger robustness.}
    \label{fig:ate_drift}
\end{figure}

We amplify the wheel odometry drift to test RF's robustness against different levels of camera motion prior drift. We also test RF's performance without the object motion prior (fix $\lambda_d=0$). The relation between the RPE of the estimated trajectories and the drift over all robot sequences is shown in \Cref{fig:ate_drift}. Even without the object motion prior, RF still achieves better performance than CF for up to 17 cm/s drift in terms of average RPE. Using both motion priors, RF performs even better and is more robust to large odometry drift. This demonstrates that our method can handle large odometry drift and the absence of an object motion prior.

\subsection{Impact of Multiple Dynamic Objects}

\begin{figure}[tbh]
    \centering
    \setlength{\tabcolsep}{0pt}
    \begin{tabular}{ccccccc}
     & \multicolumn{3}{c}{\textit{occlusion\_2\_translational}} & \multicolumn{3}{c}{\textit{swinging\_4\_translational}}
    \\
    \rotatebox[origin=c]{90}{RGB}\hspace{0.1cm}     &
    \includegraphics[width=0.157\linewidth,frame,valign=m]{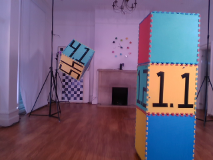}        & \includegraphics[width=0.157\linewidth,frame,valign=m]{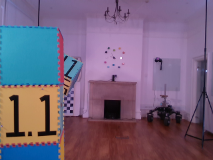}        &
    \includegraphics[width=0.157\linewidth,frame,valign=m]{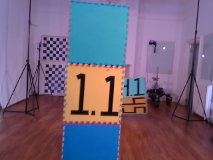}& 
    \includegraphics[width=0.157\linewidth,frame,valign=m]{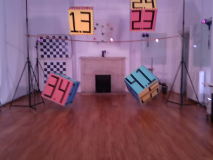}        &
    \includegraphics[width=0.157\linewidth,frame,valign=m]{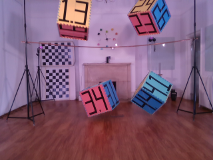}        &
    \includegraphics[width=0.157\linewidth,frame,valign=m]{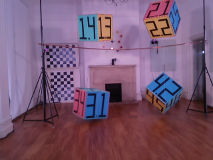}        
    \\
    \rotatebox[origin=c]{90}{SF}\hspace{0.1cm}    & \includegraphics[width=0.157\linewidth,frame,valign=m]{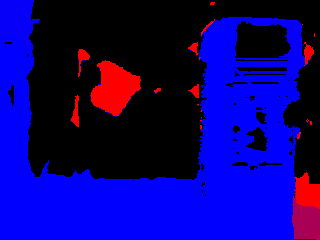} & 
    \includegraphics[width=0.157\linewidth,frame,valign=m]{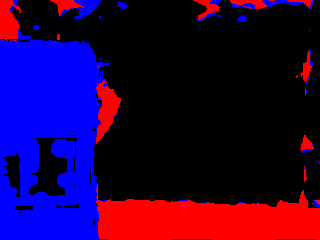} & 
    \includegraphics[width=0.157\linewidth,frame,valign=m]{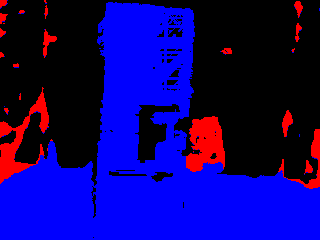} &
    \includegraphics[width=0.157\linewidth,frame,valign=m]{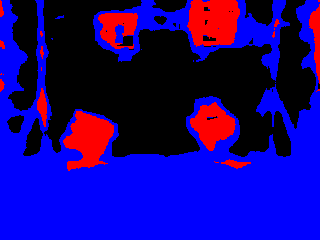} &
    \includegraphics[width=0.157\linewidth,frame,valign=m]{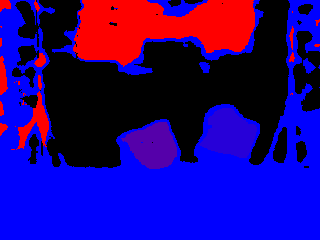} &
    \includegraphics[width=0.157\linewidth,frame,valign=m]{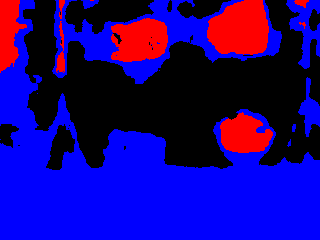}
    \\
    \rotatebox[origin=c]{90}{RF}\hspace{0.1cm}     & \includegraphics[width=0.157\linewidth,frame,valign=m]{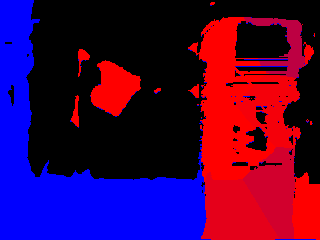}    & \includegraphics[width=0.157\linewidth,frame,valign=m]{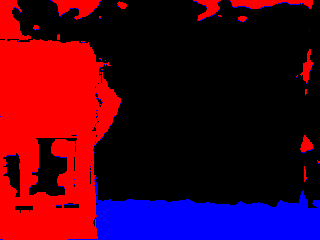}    & \includegraphics[width=0.157\linewidth,frame,valign=m]{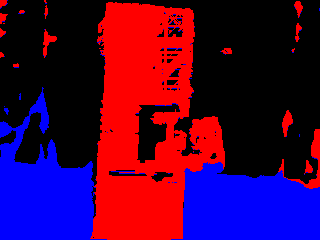}&
    \includegraphics[width=0.157\linewidth,frame,valign=m]{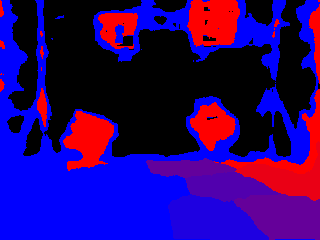} &
    \includegraphics[width=0.157\linewidth,frame,valign=m]{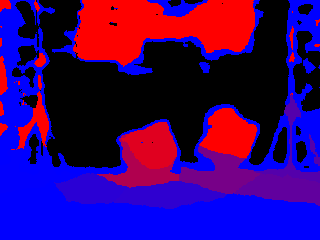} &
    \includegraphics[width=0.157\linewidth,frame,valign=m]{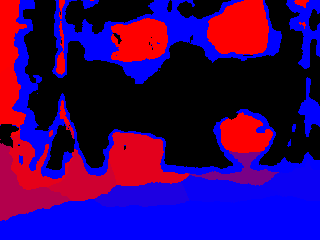}
    \end{tabular}
    \caption{Segmentation results of two OMD \cite{judd2019oxford} sequences with multiple dynamic objects. Although multiple objects can only be represented by a single transformation, RigidFusion (RF) is able to segment the static environment (blue) against multiple dynamic objects (red), while StaticFusion (SF) maps dynamic objects into the static environment.}
    \label{fig:omd}
\end{figure}

RF assumes that the dynamic motion can be explained by a single rigid transformation. To test RF's performance when this assumption is violated, we conduct qualitative experiments on two OMD sequences \cite{judd2019oxford} where multiple dynamic objects are present (\Cref{fig:omd}).

For sequence \textit{occlusion\_2\_translational}, which contains one large and one small dynamic object, the motion prior for the larger object is provided. For sequence \textit{swinging\_4\_translational}, which contains four dynamic objects, the motion prior for the top-left object is provided. Despite this under-representation of the dynamic motion, RF outperforms SF and is able to correctly segment the static environment against all the dynamic objects. However, similar to SF, RF cannot independently track multiple dynamic objects with different motions.

\section{Conclusion}
\label{sec:conclusion}
We have presented a robot localisation and mapping approach in environments where dynamic components can occupy the major portions of the visual input. To address this problem, we assume that the dynamic component is rigid, and jointly segment and track the static and dynamic rigid bodies. Detailed evaluation shows that our method RigidFusion outperforms state-of-the-art when a dynamic rigid object occludes more than 65\% of the camera view. We demonstrate its robustness to odometry drift up to 17 cm/s and the absence of object motion priors.

Our method treats the whole dynamic component as a single rigid body and is thus unable to track multiple dynamic objects independently in the scene. Our future research direction involves extending the current pipeline to enable multiple rigid object segmentation, tracking and reconstruction in dynamic environments. To handle dynamic objects that are not in contact with the manipulator, and thus have no kinematic prior, we intent to estimate motion priors using visual correspondences. 

\section*{ACKNOWLEDGEMENT}

The authors would like to thank Raluca Scona for answering questions about StaticFusion and Tin Lun Lam from AIRS for critical comments.

\bibliographystyle{IEEEtran.bst}
\bibliography{IEEEabrv,mybibfile}

\end{document}